\documentclass[lettersize,journal]{IEEEtran}
\usepackage{amsmath,amsfonts}
\usepackage{algorithmic}
\usepackage{algorithm}
\usepackage{array}
\usepackage[caption=false,font=normalsize,labelfont=sf,textfont=sf]{subfig}
\usepackage{textcomp}
\usepackage{stfloats}
\usepackage{url}
\usepackage{verbatim}
\usepackage{graphicx}
\usepackage{cite}
\usepackage{hyperref}
\usepackage{booktabs}
\usepackage{makecell} % For better cell formatting
\usepackage{scalerel}
\usepackage{amssymb}
\usepackage{pifont}
\usepackage{etoolbox}
\usepackage[table]{xcolor}

\begin{document}

%%%%% table
% \makeatletter
% \patchcmd{\@makecaption}
%   {\scshape}
%   {}
%   {}
%   {}
% \makeatother
% \renewcommand{\thetable}{\arabic{table}} %阿拉伯数字
%%%%

\title{FreeEdit: Mask-free Reference-based \\ Image Editing with Multi-modal Instruction}

% \author{IEEE Publication Technology,~\IEEEmembership{Staff,~IEEE,}
%         % <-this % stops a space
% \thanks{This paper was produced by the IEEE Publication Technology Group. They are in Piscataway, NJ.}% <-this % stops a space
% \thanks{Manuscript received April 19, 2021; revised August 16, 2021.}}
\author{Runze He,
        Kai Ma,
        Linjiang Huang,
        Shaofei Huang,
        Jialin Gao, 
        Xiaoming Wei, 
        Jiao Dai, 
        Jizhong Han, 
        Si Liu% <-this % stops a space

\IEEEcompsocitemizethanks{\IEEEcompsocthanksitem
% \begin{itemize}

% Runze He, Shaofei Huang, Jiao Dai, and Jizhong Han are with Chinese Academy of Sciences Institute of Information Engineering.

% Linjiang Huang and Si Liu are with the Institute of Artificial Intelligence, Beihang University.

% Kai Ma, Jialin Gao, and Xiaoming Wei are with Meituan.

Runze He, Shaofei Huang, Jiao Dai, and Jizhong Han are with Chinese Academy of Sciences Institute of Information Engineering. Email: \{hrz010109,nowherespyfly\}@gmail.com, \{daijiao,hanjizhong\}@iie.ac.cn.

Linjiang Huang and Si Liu are with the Institute of Artificial Intelligence, Beihang University. Email: ljhuang524@gmail.com, liusi@buaa.edu.cn.

Kai Ma, Jialin Gao, Xiaoming Wei are with Meituan, Email: \{makai20,gaojialin04,weixiaoming\}@meituan.com.
% \end{itemize}
}
}

% The paper headers
% \markboth{Journal of \LaTeX\ Class Files,~Vol.~14, No.~8, August~2021}%
% {Shell \MakeLowercase{\textit{et al.}}: A Sample Article Using IEEEtran.cls for IEEE Journals}

% \IEEEpubid{0000--0000/00\$00.00~\copyright~2021 IEEE}
% Remember, if you use this you must call \IEEEpubidadjcol in the second
% column for its text to clear the IEEEpubid mark.

\maketitle

\def\ip2p{InstructPix2Pix}
\def\blipdf{BLIP-Diffusion}
\def\ours{FreeEdit}
\def\ourdata{FreeBench}
\def\mimic{MimicBrush}
\def\db{DreamBooth}
\def\qf{Q-Former}
\def\cd{CustomDiffusion}
\def\cog{CogVLM}
\def\mini{Mini-CPM}
\def\llava{LLaVA}
\def\alphaclip{AlphaCLIP}
\def\p2p{Prompt2Prompt}
\def\ptp{Prompt2Prompt}
\def\ti{Textual Inversion}
\def\refnet{ReferenceNet}
\def\masactrl{MasaCtrl}
\def\paintby{PaintByExample}
\def\anydoor{AnyDoor}
\def\kosmosg{Kosmos-G}
\def\magicbrush{MagicBrush}
\def\ipadapter{IP-Adapter}
\def\emuedit{Emu Edit}
\def\replace{REPLACE}
\def\add{ADD}
\def\remove{REMOVE}
\def\style{STYLE}
\def\design{DesignEdit}
\def\openimage{OpenImage}
\def\dreamedit{DreamEdit}
\def\dragondiff{DragonDiffusion}
\def\elite{ELITE}
\def\pp{PowerPaint}
\def\bn{BrushNet}
\def\ctn{ControlNet}
\def\clip{CLIP}
\def\unet{U-Net}
\def\mm{multi-modal}
\def\mb{MagicBrush}

\def\dui{\scalerel*{\includegraphics{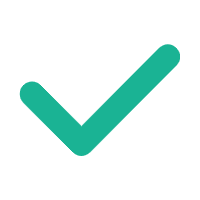}}{\textrm{\textbigcircle}}}

\def\cuo{\scalerel*{\includegraphics{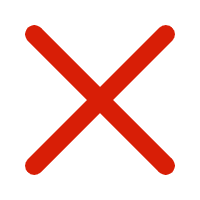}}{\textrm{\textbigcircle}}}

\begin{abstract}

Introducing user-specified visual concepts in image editing is highly practical as these concepts convey the user's intent more precisely than text-based descriptions.
We propose \ours{}, a novel approach for achieving such reference-based image editing, which can accurately reproduce the visual concept from the reference image based on user-friendly language instructions.
Our approach leverages the \mm{} instruction encoder to encode language instructions to guide the editing process. This implicit way of locating the editing area eliminates the need for manual editing masks.
To enhance the reconstruction of reference details, we introduce the Decoupled Residual Refer-Attention (DRRA) module. This module is designed to integrate fine-grained reference features extracted by a detail extractor into the image editing process in a residual way without interfering with the original self-attention.
Given that existing datasets are unsuitable for reference-based image editing tasks, particularly due to the difficulty in constructing image triplets that include a reference image, we curate a high-quality dataset, \ourdata{}, using a newly developed twice-repainting scheme. \ourdata{} comprises the images before and after editing, detailed editing instructions, as well as a reference image that maintains the identity of the edited object, encompassing tasks such as object addition, replacement, and deletion.
By conducting phased training on \ourdata{} followed by quality tuning, \ours{} achieves high-quality zero-shot editing through convenient language instructions. We conduct extensive experiments to evaluate the effectiveness of \ours{} across multiple task types, demonstrating its superiority over existing methods.
The code will be available at: \url{https://freeedit.github.io/}.

\end{abstract}


\section{Introduction}

\IEEEPARstart{W}{ith} the rise of social media, the demand for individual users to create images has grown significantly. Traditional image creation relies on professional tools and labor-intensive processes. The emergence of artificial intelligence has lowered the technical barriers, enabling amateur users to generate images more freely and easily.

In this context, diffusion models~\cite{Ho2020DDPM,Song2021DDIM} have rapidly advanced, offering more stable training and the ability to generate high-quality, diverse images. These models are gradually replacing GANs~\cite{goodfellow2020generative,karras2019style,zhang2022styleswin} as the leading technique in image synthesis. Trained on large-scale datasets, diffusion models~\cite{Ramesh2021DALLE,Ramesh2022DALLE2,Saharia2022Imagen,Rombach2022StableDiffusion,GLIDE} have demonstrated powerful capabilities in synthesizing high-fidelity images from text prompts. Beyond text-based image generation, diffusion models are widely utilized for tasks such as generating images from various condition signals like sketches and depth maps~\cite{zhang2023ctn,zhao2024unicnt}, image inpainting~\cite{mimicbrush,zhuang2023powerpaint,lugmayr2022repaint,xie2023smartbrush}, image stylization~\cite{zhang2023inversion,huang2024diffstyler,wang2024instantstyle1,wang2024instantstyle2}, and more.

\begin{figure}[!t]
\centering
\includegraphics[width=3.5in]{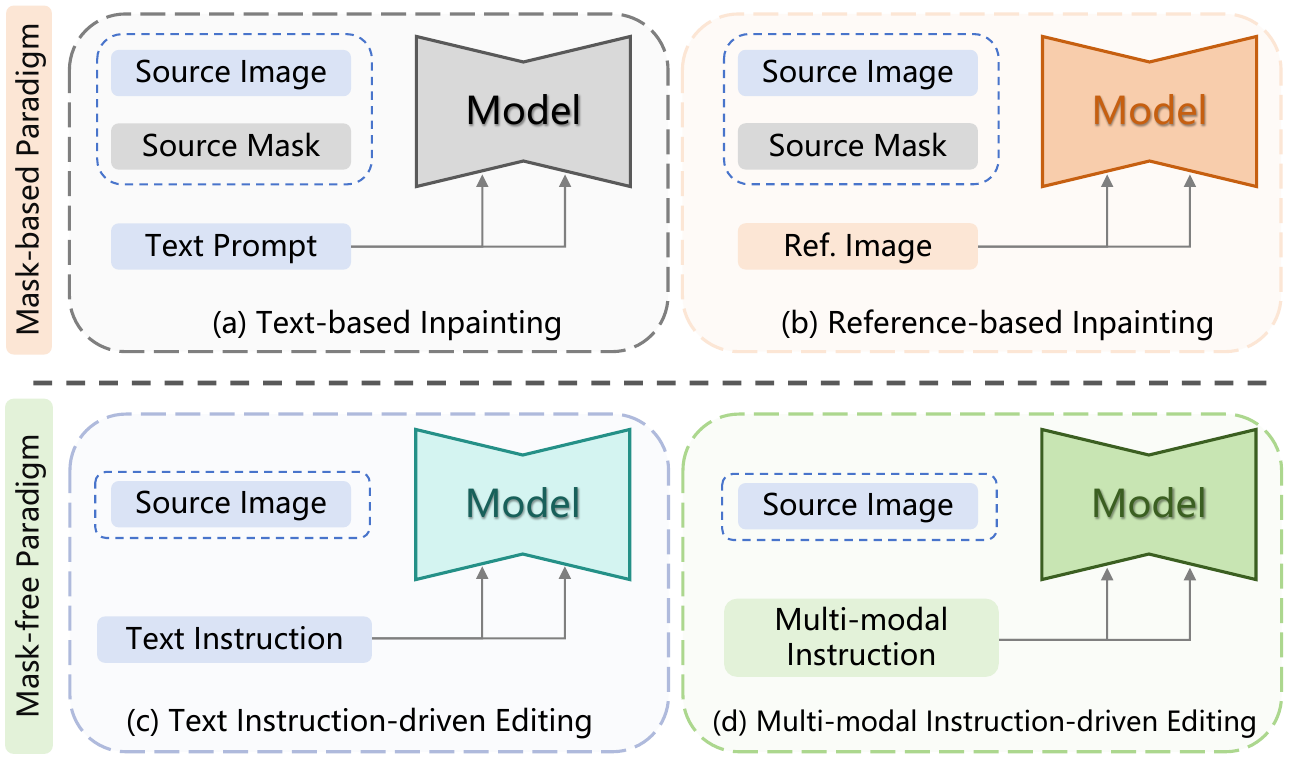}
\caption{Comparison between mask-based paradigm and mask-free paradigm, the former requires the user to provide the source mask to specify the editing area, while the latter conditions the diffusion model with language instructions, without the need for the masks. Reference-based inpainting conditions the model on reference image embedding, no longer supporting natural language. We use multi-modal instruction to introduce reference image features while still retaining the perception of natural language.}
\label{fig:task}

\end{figure}

Among these tasks, image editing has also become widely adopted using diffusion models. Recently, instruction-driven image editing~\cite{brooks2023instructpix2pix,Zhang2023MagicBrush,Sheynin2023Emu_edit} has gained significant attention, as it allows users to edit images using text instructions, aligning more closely with natural user habits. 
However, textual descriptions cannot accurately convey the user's editing intent, even if it's very detailed as discussed in previous works~\cite{yang2022paintby,gal2022ti}.
% With the growing demand for user-preference image editing, reference-based image editing has recently gained significant attention.
With the growing demand for user-preference image editing, reference-based image editing is being widely explored.
Given a reference image, reference-based image editing aims to modify the target image by incorporating elements from the reference while maintaining overall coherence and quality.
% \textcolor{blue}{or styles}
Nevertheless, the implementation of this task is non-trivial, which entails achieving two primary objectives: locating the editing region and introducing the reference appearance.
% \textcolor{blue}{separate current solutions into two parts: 1. to locate the editing region, several methods try to use user-provided masks, ..., its limitation. 2. using pre-trained encoders, its drawback.}
{Mask-based methods~\cite{chen2023anydoor,yang2022paintby,mimicbrush} introduce manual masks to locate the editing region to simplify the problem. This reliance on masks causes them to lose the flexibility to use language instructions to control editing. To introduce the reference appearance, previous methods~\cite{chen2023anydoor,yang2022paintby} typically utilize pre-trained image encoders~\cite{Radford2021CLIP,Oquab2023DINOv2,Caron2021DINO}. However, these encoders often fail to fully capture the intricate details within the reference images, leading to insufficient similarity between the generated output and the reference images.}

To tackle the aforementioned challenges, we propose \ours{}, an innovative mask-free reference-based image editing framework with precise reference appearance injection.
As shown in Figure~\ref{fig:task}, our approach differs from previous inpainting-based methods by conditioning the diffusion model on image-text \mm{} instructions rather than abstract image embeddings, eliminating the need to manually specify an editing area.
% \textcolor{red}{why?}
To encode user-provided language instructions, \ours{} elaborates a \mm{} instruction encoder that transforms them into editing instruction embeddings.
These embeddings are then incorporated into the diffusion model via the cross-attention mechanism, guiding the editing process effectively.
To address the limitations of existing methods in preserving the details of reference subjects, we further introduce the Decoupled Residual Refer-Attention (DRRA) module.
Unlike the previous methods~\cite{hu2024animate,chen2024magiccloth,mimicbrush}, which directly integrates the reference attention into the self-attention, DRRA separates reference attention from self-attention and learns it residually, maintaining the integrity of the original self-attention.
By extracting fine-grained features from reference images, DRRA ensures the identity (ID) consistency between the edited result and the reference subject.
To effectively introduce the reference information, we leverage a phased training process including instruction embedding and fine-grained feature integration. Finally, a quality tuning stage enhances the overall editing quality, ensuring high fidelity and consistency.

To address the lack of datasets for reference-based instruction-driven editing, we developed \ourdata{}, a high-quality dataset designed to support model training. Previous studies always struggled with constructing image triplets that include the original, edited, and reference images, because it is difficult to maintain the ID consistency of the reference images and the edited images.
% \textcolor{blue}{Reason?}
% To resolve this, we build \ourdata{} upon a real-world segmentation dataset, \textcolor{red}{claim the reason.}
% \textcolor{blue}{One sentence to ensure coherence.} 
% \textcolor{blue}{details}
To resolve this, we implement a twice-repainting construction scheme to ensure identity consistency between the edited object and the reference, based on the real-world segmentation dataset.
To further enhance the model's ability to comprehend language instructions, we perform instruction recaptioning on the constructed image triplets, providing as detailed and accurate instructions as possible. After meticulous data construction and filtering, we finally obtain a dataset containing 131,160 edits. Different from previous datasets~\cite{brooks2023instructpix2pix,Zhang2023MagicBrush}, the instructions in \ourdata{} are multi-modal, combining text and reference images to cover tasks such as object addition, replacement, and removal.

Extensive experiments demonstrate that \ours{} surpasses previous methods in reference-based image editing tasks, such as object addition and replacement, without requiring editing masks. It also achieves comparable results in object removal and plain-text instruction-driven editing tasks. Our contributions can be summarized as follows:
\begin{itemize}
\item We are the first to implement reference-based image editing in an instruction-driven manner, eliminating the need for manual masks and significantly improving ease of use.
\item We propose a powerful editing model, \ours{}, which utilizes \mm{} instructions to implicitly locate editing regions and incorporates DRRA modules to efficiently integrate fine-grained reference image features.
\item We introduce a high-quality dataset, \ourdata{}, specifically designed for reference-based image editing tasks, containing a total of 131,160 edits to support and foster research in the community. 
\end{itemize}

\section{Related Work}

\subsection{Image Customization}

To perform image generation conditioned on a given subject, image customization methods~\cite{ruiz2022dreambooth,gal2022ti,kumari2022customdiffusion} that require fine-tuning are developed to learn specific visual concepts in a given image set, which attract a lot of attention from the community and have made great progress. Given 3-5 reference images for a targeted subject, \db{} \cite{ruiz2022dreambooth} enables the model to learn subject-specific concepts by fine-tuning on the given small-scale image set with a class-speciﬁc prior preservation loss to keep its generation ability. \ti{} \cite{gal2022ti} learns the concept of a new subject by optimizing a new token in the text space. \cd{}, similar to \ti{}, learns a new token in the text space and fine-tunes a few parameters in the text-to-image conditioning mechanism to achieve better generation quality and faster fine-tuning.

Compared to the above methods which need fine-tuning for each subject, zero-shot image customization methods~\cite{wei2023elite, li2023blipdf, ye2023ip-adapter} are more appealing. \elite{} \cite{wei2023elite} encodes visual features from the CLIP image encoder into textual embeddings and introduces patch features into cross-attention layers to provide details. \blipdf{} \cite{li2023blipdf} takes the \qf{} to extract image features, yielding promising image customization results after training on a large amount of data. IP-Adapter \cite{ye2023ip-adapter} injects image embeddings using a decoupled cross-attention mechanism to achieve text-compatible image prompts. However, these methods are unsuitable for reference-based image editing due to significant differences in tasks.

\subsection{Image Inpainting}
Image inpainting is closely related to the task of image editing. Traditional image inpainting methods~\cite{li2022mat, lugmayr2022repaint} perform unconditional completion given masks, while recent advancements in text-based image inpainting have continually improved this task. SD-Inpainting modifies the input channels of the diffusion model to 9 channels, allowing for additional mask and masked image conditions. SmartBrush~\cite{xie2023smartbrush} enables text and shape-guided object inpainting by introducing object-mask prediction. \pp{}~\cite{zhuang2023powerpaint} achieves high-quality versatile image inpainting with learnable task embeddings. \bn{}~\cite{ju2024brushnet} designs decomposed dual-branch diffusion to serve as a plug-and-play solution for image inpainting.

Text-based inpainting, however, cannot fully capture the user's intent when it comes to specific visual concepts. To address this, \paintby{}~\cite{yang2022paintby} replaces the text embedding condition with the image embedding obtained from the CLIP image encoder~\cite{Radford2021CLIP} to achieve exemplar-based image inpainting. \anydoor{}~\cite{chen2023anydoor} introduces a powerful DINO image encoder~\cite{Caron2021DINO,Oquab2023DINOv2} and the high-frequency map to promote the similarity of the results to the reference images. Our contemporaneous method, \mimic{}~\cite{mimicbrush}, incorporates reference imitation to perform imitative editing, and it still follows the inpainting paradigm. As an implementation of reference-based image editing, the inpainting-based methods present promising editing results, but they require users to provide editing areas, and cannot condition the diffusion model by convenient language instructions.

\subsection{Image Editing}

Image editing methods~\cite{Hertz2022Prompt2prompt, mokady2023null, Tumanyan_2023_CVPR_pnp, cao_2023_masactrl, Imagic} emerges in large numbers recently. \p2p{} \cite{Hertz2022Prompt2prompt} performs zero-shot image editing using a pair of text descriptions through cross-attention manipulation. NullTextInversion~\cite{mokady2023null} optimizes the unconditional textual embeddings to promote the editing of real images based on \ip2p{}. PnP~\cite{Tumanyan_2023_CVPR_pnp} manipulates spatial features and their self-attention inside the model to facilitate fine-grained control. MasaCtrl~\cite{cao_2023_masactrl} introduces mutual self-attention for consistent image editing.
Imagic~\cite{Imagic} produces a text embedding that aligns with both the input image and the target text, and fine-tunes the diffusion model to allow high-quality complex semantic image editing.

In addition to above description-based editing methods, instruction-driven image editing ones~\cite{brooks2023instructpix2pix, Zhang2023MagicBrush, Sheynin2023Emu_edit, fu2024mgie, huang2023smartedit, geng2023instructdiffusion} have attracted more and more attention from the community, as text instructions are more in line with human habits of image editing. Early work \ip2p{} \cite{brooks2023instructpix2pix} achieves the task by fine-tuning the diffusion model on the constructed dataset which is constructed with LLMs~\cite{achiam2023gpt,touvron2023llama2,Openai2022ChatGPT} generating text pairs and \ptp{}~\cite{Hertz2022Prompt2prompt} generating image pairs then. MagicBrush~\cite{Zhang2023MagicBrush} presents a human-annotated high-quality instruction-driven editing dataset. \emuedit{}~\cite{Sheynin2023Emu_edit} builds on Emu \cite{dai2023emu} by scaling up dataset construction and integrating traditional computer vision tasks, akin to InstructDiffusion~\cite{geng2023instructdiffusion}. While these methods have made image editing more accessible, they do not yet support reference-based image editing tasks.

\section{Approach}

In this section, we begin with an overview of \ours{} in Section~\ref{method:over}. Next, Section~\ref{method:mm_ins} introduces our \mm{} editing instruction. We then provide a detailed description of the fine-grained feature injection in Section~\ref{method:kv}. Finally, Section~\ref{method:infer} covers our training strategies.

\begin{figure*}[!t]
\centering
\includegraphics[width=7.2in]{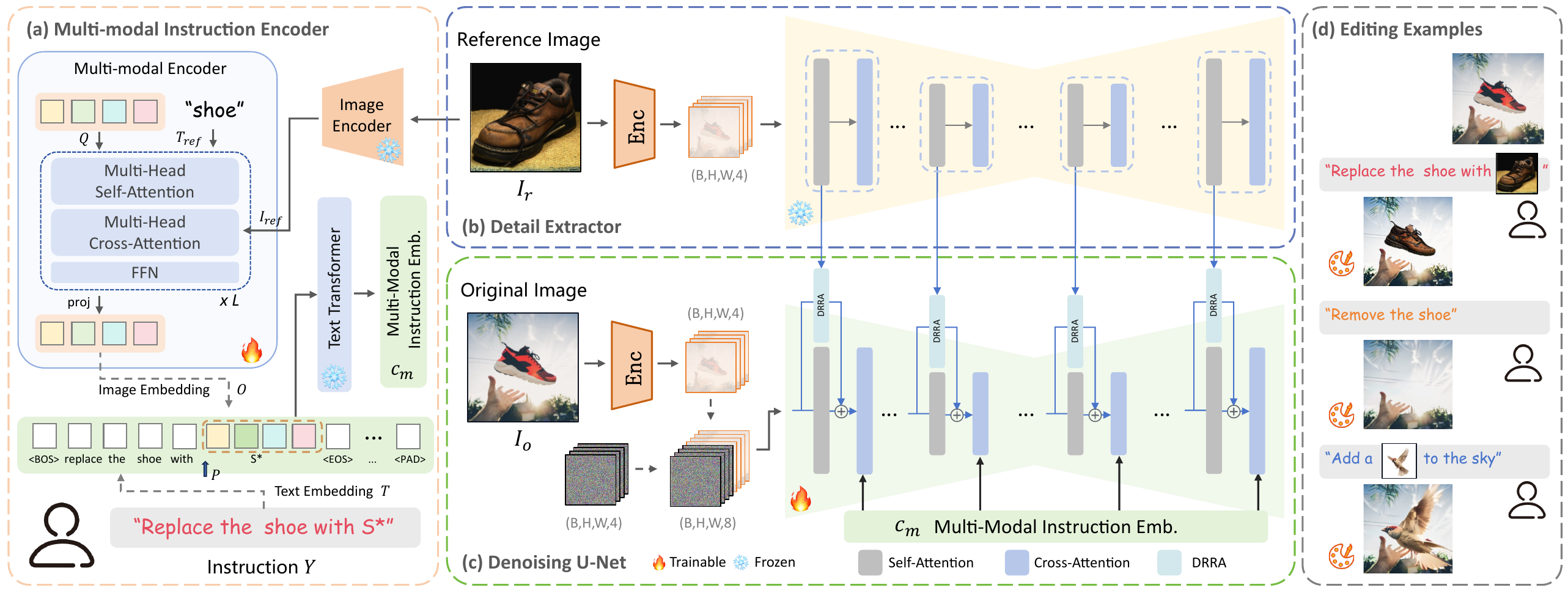}
\caption{The overall pipeline of our proposed \ours{}, which consists of three components: (a) Multi-modal instruction encoder. (b) Detail extractor. (c) Denosing \unet{}. Text instruction and reference image are firstly fed into the multi-modal instruction encoder to generate \mm{} instruction embedding. The reference image is additionally fed into the detail extractor to obtain fine-grained features. The original image latent is concatenated with the noise latent to introduce the original image condition. Denosing \unet{} accepts the 8-channel input and interacts with the multi-modal instruction embedding through cross-attention. The DRRA modules which connect the detail extractor and the denoising \unet{}, are used to integrate fine-grained features from the detail extractor to promote ID consistency with the reference image. (d) The editing examples obtained using \ours{}.}
\label{fig:pipe}

\end{figure*}

\subsection{Overview}
\label{method:over}
The overall pipeline of \ours{} is represented in Figure~\ref{fig:pipe}. Given an original image $I_o$ and a language instruction $Y$ with a reference image $I_r$, \ours{} edits image $I_o$ according to the intent of the language instruction, and the edited image maintains the ID consistency with the visual concept in $I_r$.
% For example, we can use the instruction “replace the shoe with S*” to replace the red-black shoe in the original image with a specific shoe in the reference image, “remove the shoe” to remove the shoe from the image, and “add a S* to the sky” to add a specific bird in the sky, where S* represents the visual concept from the reference image.

\ours{} consists of a \mm{} instruction encoder, a detail extractor and a denosing \unet{}. The \mm{} instruction encoder encodes the language instructions into multi-modal editing instruction embedding to condition the denoising \unet{}. The detail extractor shares the same architecture as \unet{}, which extracts multi-scale fine-grained features from the reference image. The denoising \unet{} accepts 8-channel latent input for image editing, with the DRRA modules integrating the reference features obtained from the detail extractor to promote the ID consistency of the editing result with the reference image.

\subsection{Multi-modal Editing Instruction}
\label{method:mm_ins}
Previous reference-based image editing methods~\cite{chen2023anydoor,yang2022paintby,mimicbrush} condition the diffusion model on the reference image by substituting the text embedding with the projected reference image embedding.
% This practice, however, results in a dependency on manual masking because models with image embeddings as the cross-attention control condition no longer have text localization capabilities.
However, this practice leads to a reliance on manual masking, as models utilizing image embeddings for the cross-attention mechanism lose their ability to perform editing region localization by texts.
% \textcolor{blue}{why?}
% restricts the model's ability to utilize texts to localize the editing area, \textcolor{blue}{why?}.
% This practice, however, restricts the model's ability to utilize texts to localize the editing area, resulting in a dependency on manual masking \textcolor{blue}{why?}.
% As a result, only extra input masks can be utilized
% requiring an additional mask to address this limitation
Furthermore, the trained model is only specialized for the reference-based inpainting task, which means it lacks the flexibility to perform object removal or other image editing operations that are typically guided by plain text instructions.
Instead, we seek to implement reference-based image editing in a mask-free manner.
Achieving such a goal is challenging, compared to those methods which are given the mask of editing region beforehand. 
Intuitively, one promising approach is to employ text-image interlaced editing instructions, exemplified by commands like ``\emph{replace the bird in the sky with \textbf{S*}}''. This multi-modal instruction not only implicitly defines the editing area as ``\emph{the bird in the sky}'', but also incorporates visual cues from the reference image, denoted as \textbf{S*}. Moreover, the feature space of these multi-modal instructions closely aligns with the text feature space, enabling the development of a versatile editing model that can also accommodate tasks such as object removal.

To encode the multi-modal instructions, it is essential to extract visual features from the subjects within the reference images.
We opt for the multi-modal encoder \qf{}, which is prevalent in MLLMs~\cite{zhu2023minigpt,li2022blip}.
In contrast to single-modal image encoders~\cite{Radford2021CLIP,Caron2021DINO,Oquab2023DINOv2}, it also allows interaction with the reference text features through additional attention modules.
This advantage enables us to leverage corresponding reference texts to mitigate the impact of background noise or other visual elements, thereby focusing on the extraction of subject features.

Specifically, $N$ learnable queries $Q=[q_1,q_2,...,q_N]$ and reference text features $T_{ref}$ are concatenated together to consider the reference text conditions,

\begin{equation}
\begin{aligned}
% \label{eq:rasa}
W_0=[Q,T_{ref}],
\end{aligned}
\end{equation}

\noindent which are further fed into $L$ stacked attention modules in \qf{}. In each attention module, features are aggregated by self-attention and interact with reference image features $I_{ref}$ through cross-attention. The process can be formulated as:
% \textcolor{blue}{better with an equation}

\begin{equation}
\begin{aligned}
% \label{eq:rasa}
W_{l}=\texttt{FFN}(\texttt{MHCA}(\texttt{MHSA}(W_{l-1}), I_{ref})),\ \mathrm{for} \ l=1,\dots,L,
\end{aligned}
\end{equation}

\noindent where \texttt{MHCA}, \texttt{MHSA} denote multi-head cross-attention and multi-head self-attention respectively, and \texttt{FFN} denotes feed-forward layer. We obtain final subject visual features $O=[o_1,o_2,...,o_N]$ through an extra projection layer, formally,

\begin{equation}
\begin{aligned}
% \label{eq:rasa}
O= \texttt{proj}(W_L).
\end{aligned}
\end{equation}

We can think of the updated queries $O$ as pseudo-word vectors that represent specific visual concepts, and insert them into the corresponding positions $P$ in the text word vectors $T$. After the interaction in the frozen CLIP text encoder, the final multi-modal instruction embedding $c_m$ is generated, formally:
% \textcolor{blue}{Please show $c_m$ in the figure. What does the white rectangle after image embeddings stand for? $<$ EOS $>$ token? Annotate it or remove it, aligning with your instruction.}.

\begin{equation}
\begin{aligned}
% \label{eq:rasa}
\small
% c_m=\mathrm{\Phi}({T},{O},{P})
c_m=\texttt{TextEncoder}({T},{O},{P}).
\end{aligned}
\end{equation}

\subsection{Fine-grained Feature Injection}
\label{method:kv}

The pre-trained image encoder may not capture every nuance of the reference subject, so relying solely on the multi-modal instruction embedding is insufficient for reconstructing the high-fidelity details of the reference subject, which is a challenge also encountered by prior work utilizing image embedding conditioned diffusion models. 
To solve this problem, we design the fine-grained feature injection to improve the ID consistency between the editing result and the reference subject.

\begin{figure}[!t]
\centering
\includegraphics[width=3.5in]{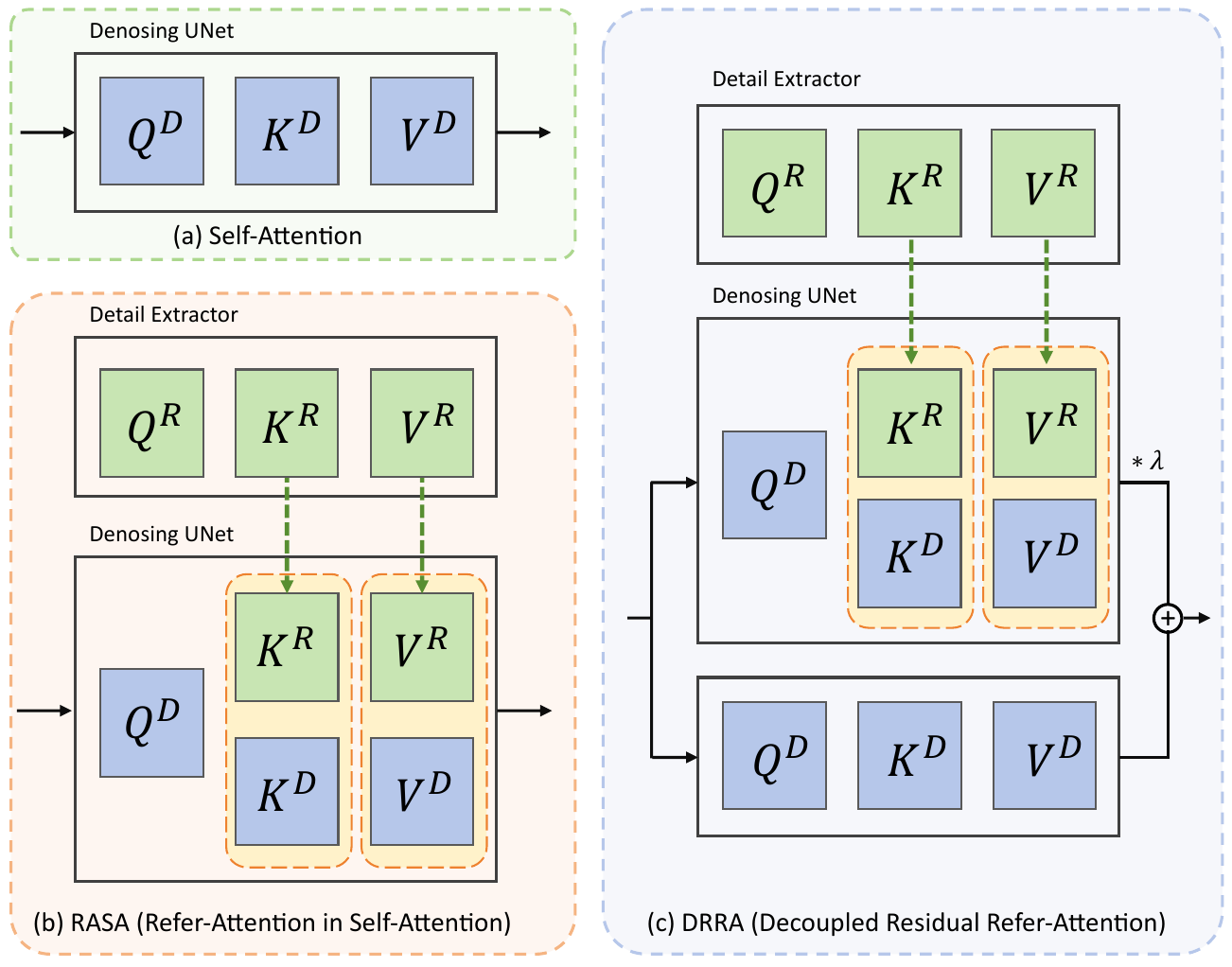}
\caption{Comparison between (a) Self-Attention (b) Refer-Attention in Self-Attention (RASA) and (c) Decoupled Residual Refer-Attention (DRRA). RASA performs additional attention to reference features obtained from the detail extractor by concatenating them to the original self-attention module. DRRA retains the original self-attention and implements the decoupled reference attention in the form of residual connection.}
% \textcolor{blue}{These notations should be presented in text, better using equations.}
\label{fig:refer_attn}

\end{figure}

As explored in previous approaches~\cite{chen2024magiccloth,hu2024animate}, the pre-trained image generation model is a good image feature extractor in its own right that can be used for a variety of downstream tasks. To fully consider the visual features of the reference image in the editing process, we introduce a detail extractor that mirrors the architecture of \unet{} as an image feature encoder to obtain the multi-scale fine-grained features of the reference image.
Specifically, we extract the features from the self-attention modules, which contain a wealth of information. Since the features of the detail extractor have the same scale and are in the same feature space with denoising \unet{}, they can be easily and efficiently used to guide the image editing process.

Having acquired the fine-grained features through the detail extractor, the subsequent critical step is the seamless integration of these features into the image-denoising process. This integration aims to enhance the resemblance between the edited output and the reference subjects without damaging the editing ability of the original diffusion model.
As shown in Figure~\ref{fig:refer_attn}, the previous methods enable reference features injection by expanding original self-attention into refer-attention, that is, the reference features are concatenated to the key and value in the self-attention layers. When the model performs self-attention, it will also take into account the reference features, which we call Refer-Attention in Self-Attention (RASA).
Formally, given the attention features $Q^R$, $K^R$, $V^R$ from the self-attention module of the detail extractor and the attention features $Q^D$, $K^D$, $V^D$ from the self-attention module of the denoising \unet{}, the resulting output from this module is defined by:

% Formally, given the reference feature $f_r \in R_{b \times (h \times w) \times d}$ from the detail extractor and the base feature $f \in R_{b \times (h \times w) \times d}$ from denoising \unet{}, where $b$, $h$, $w$, and $d$ refer to the batch size, height, width, and channel dimensions respectively, the resulting output from this module is defined by:

% \begin{equation}
% \begin{aligned}
% \label{eq:rasa}
% \small
% \mathrm{RASA}(f,f_r)=\mathrm{Softmax}(\frac{\mathbf{W}_Qf(\mathbf{W}_K[f,f_r])^T}{\sqrt{d}})\mathbf{W}_V[f,f_r]
% \end{aligned}
% \end{equation}

\begin{equation}
\begin{aligned}
% \label{eq:rasa}
\small
% \mathbf{W}_K
\texttt{Out}=\texttt{Softmax}(\frac{Q^D([K^D,K^R])^T}{\sqrt{d}})[V^D,V^R],
\end{aligned}
\end{equation}

% Where $W_Q$, $W_K$, and $W_V$ are the projection matrices of query, key, and value in the attention module.
\noindent where $d$ is the dimension of the attention features. This way of introducing reference features is not good enough, which is reflected in the following two aspects:
firstly, the original self-attention module is modified to additionally handle attention to reference features, resulting in a coupling of self-attention and refer-attention.
This coupling may not be favorable for conditioning the intensity of reference attention, and it might interfere with original self-attention during training, potentially leading to sub-optimal results.
Secondly, the fine-tuned model, modified in this manner, exhibits a lack of flexibility, because the updated self-attention module may become specialized to such tasks and no longer support the non-reference task well.

To deal with the above problems, we propose Decoupled Residual Refer-Attention (DRRA) to replace RASA, which decouples the refer-attention from the original self-attention and acts as a plug-and-play residual module that supports the degree adjustment of the attention paid to the reference features. Figure~\ref{fig:refer_attn} illustrates the difference between original self-attention, RASA, and DRRA. Specifically, we keep the original self-attention module unchanged and insert a specially implemented reference attention module in parallel. We introduce hyperparameter $\lambda$ to control the degree of attention to the reference features, which defaults to 1.
The modulated output from the refer-attention module is then combined with the output from the self-attention module through a residual connection, defined formally as follows:

% \begin{equation}
% \begin{aligned}
% \label{eq:drra}
% \small
% \mathrm{DRRA}(f,f_r)=\mathrm{Self\text{-}Attention}(f)+\lambda*\mathrm{RASA}(f,f_r)
% % \mathrm{DRRA}(f,f_r)=\mathrm{Self\text{-}Attention}(Q^D,K^D,V^D)+\lambda*\mathrm{RASA}(Q^D,K^D,V^D,Q^R,K^R,V^R)
% \end{aligned}
% \end{equation}

\begin{equation}
\begin{aligned}
\texttt{Out}&= \texttt{Softmax}\left(\frac{Q^D (K^D)^T}{\sqrt{d}}\right)V^D \\
&+\lambda \cdot \texttt{Softmax}\left(\frac{Q^D ([K^D, K^R])^T}{\sqrt{d}}\right)[V^D, V^R].
\end{aligned}
\end{equation}

Since the newly introduced reference attention module and the original self-attention module are separated, this allows reference attention to focus on introducing details from the reference image, while self-attention retains its original functionality without being disturbed. To minimize the interference with the original model architecture, we also zero-initialize the output of the newly inserted reference attention module. This initialization ensures a smooth transition and minimal interference with the existing model's performance.

\subsection{Training Strategies}
\label{method:infer}

\noindent \textbf{Phased Training.}
% \textcolor{blue}{Why? Explain your motivation, not just the description.}
Considering the slight differences between the training targets of the multi-modal editing instruction and the fine-grained feature injection,  i.e., the former is mainly used for localization and coarse editing, and the latter is used to supplement the details from the reference image, we divide the training of \ours{} into the multi-modal instruction training phase and the fine-grained feature injection phase.
This allows the model to focus on learning localization and coarse editing in the first stage, and in the second stage, it could focus on learning the details from the reference images, simplifying and stabilizing the training process.

In the multi-modal instruction training phase, the parameters of the \qf{} and \unet{} are fine-tuned to learn editing under the condition of multi-modal instructions. Given noisy image $x_t$ at timestep $t$, the model learns to predict noise $\epsilon$ under the condition of the original image latent $x_o$ and the \mm{} editing instruction embedding $c_m$.
The training objective of the editing model in this stage can be denoted as:

\begin{equation}
\label{eq:ldm}
% \small
\mathbb{E}_{\epsilon,\mathbf{x},\mathbf{x_o},\mathbf{c_m},t}[w_t||\epsilon-\epsilon_\theta(\mathbf{x}_t,\mathbf{x}_o,\mathbf{c_m},t)||],
% \mathcal{L}_{LDM}=\mathbb{E}_{\mathbf{z} \sim \mathcal{E}(\mathbf{x}),\mathcal{E}(\mathbf{x_{I}}),\epsilon\sim\mathcal{N}(0,1),t,c_m}\lbrack\lVert\epsilon_\phi(\mathbf{z}_t;t,\mathcal{E}(\mathbf{x_{I}}),c)-\epsilon \rVert_2^2\rbrack,
\end{equation}

\noindent where $\epsilon_{\theta}$ is the noise predictor and $w_t$ is a time-dependent weight.

In the fine-grained feature injection stage, only the parameters of the refer-attention module in DRRA are fine-tuned to introduce detailed reference features $f_{ref}$ into the editing process, and the noise predictor is further strengthened as $\epsilon_\theta(\mathbf{x}_t,\mathbf{x}_o,\mathbf{c_m},{f_{ref}},t)$. The total number of trainable parameters in this stage is only 47.28MB.

\noindent \textbf{Quality Tuning.}
Inspired by recent advanced image generation works~\cite{chen2023pixartalpha, kolors}, which perform a quality improvement stage with specifically curated high-aesthetic data following extensive training, we adopt a similar approach to fine-tune the editing model.
Specifically, we select 5K high-quality editing cases from \ourdata{}. This dataset, together with \mb{} dataset, is used as training data for the quality tuning, considering that the \mb{} is a high-quality human-annotated plain-text instruction-driven editing dataset.
At this stage, the parameters of \qf{}, \unet{}, and DRRA are all updated. The quality tuning allows the final editing model to handle both reference-based and text-only image editing tasks and deliver higher-quality editing performance.

\section{Dataset}

\begin{figure*}[!t]
\centering
\includegraphics[width=7.1in]{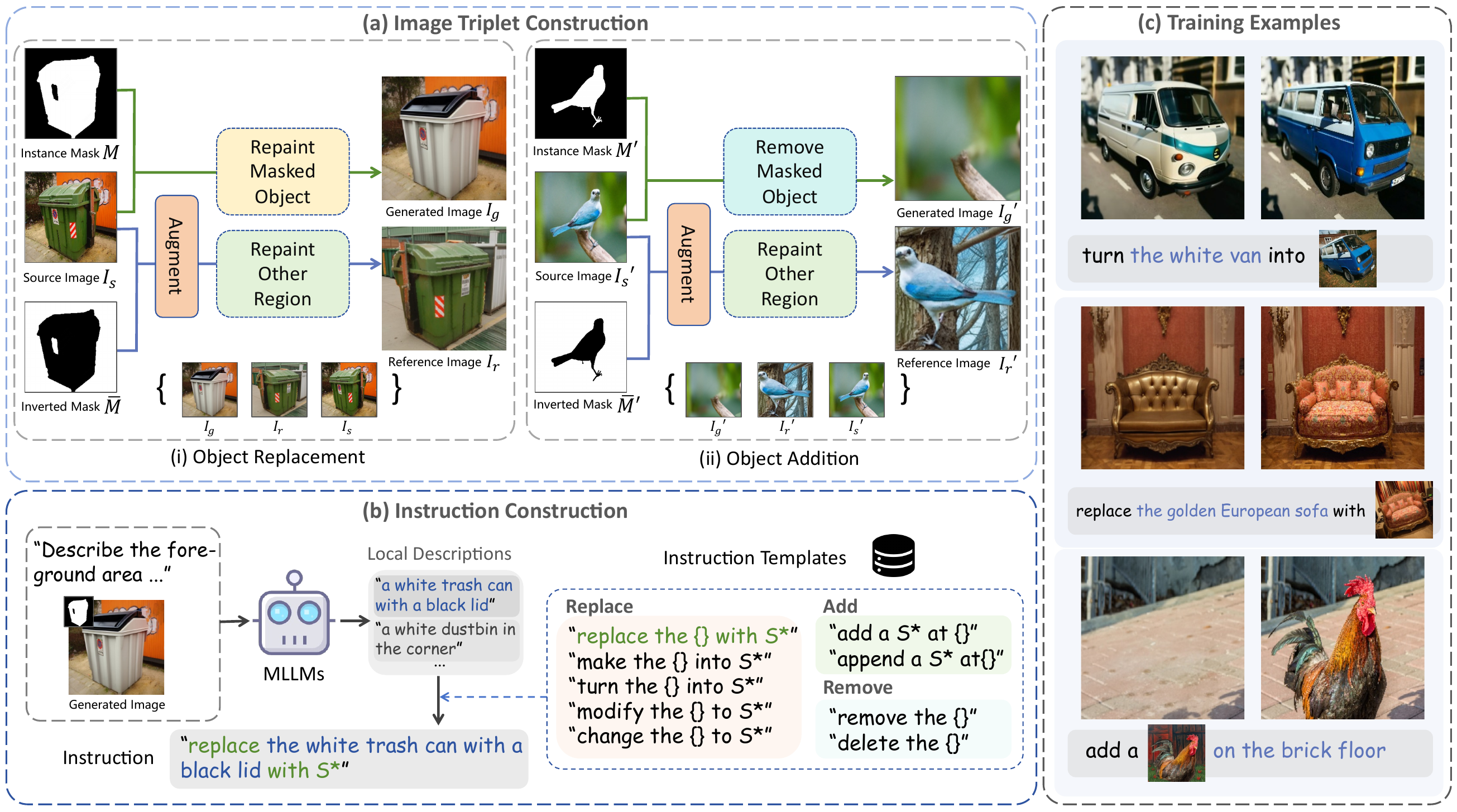}
\caption{Pipeline for dataset construction and examples of training samples. (a) Image triplet construction. We repaint the source image in the existing real-world segmentation dataset twice to form the image triplet. (b) Instruction Construction. We use multiple powerful MLLMs to caption the generated image, and combine the resulting local descriptions with instruction templates to form edit instructions. (c) Examples of the training dataset. The item in the dataset contains images before and after editing and a multi-modal instruction.}
\label{fig:data}

\end{figure*}

In this section, we delve into the details of our proposed dataset, denoted as \ourdata{}. In Section~\ref{data:image}, we describe the construction of the image triplet, which consists of the original image, the edited image, and the corresponding reference image. Subsequently, in Section~\ref{data:ins}, we introduce the method for obtaining detailed textual editing instructions. Lastly, Section~\ref{data:filter} elaborates on our data filtering process.

\subsection{Image Triplet Construction}
\label{data:image}

The previous instruction-driven image editing methods~\cite{brooks2023instructpix2pix,Sheynin2023Emu_edit} leverage the in-context learning capabilities of Large Language Models (LLMs)~\cite{Openai2022ChatGPT,touvron2023llama2} to generate extensive pairs of image descriptions and editing instructions. Subsequently, they employ the \ptp{}~\cite{Hertz2022Prompt2prompt} to produce corresponding image pairs.
However, it is not well-suited for reference-based image editing tasks.
% While this data construction pipeline facilitates instruction-driven image editing, it is not well-suited for reference-based image editing tasks.
The challenge lies in simultaneously generating a reference image that maintains ID consistency with the corresponding object in the edited image.
Another straightforward solution is to apply existing reference-based inpainting techniques~\cite{yang2022paintby,mimicbrush,chen2023anydoor}, which are to generate an edited image using an original image and a reference image.
However, such a solution limits the ID similarity between the edited result and the reference image, as these techniques fail to reconstruct the intricate details of the reference object accurately.
% Furthermore, since these image editing pairs are both generated, models trained on these data could lead to sub-optimal outcomes when faced with real image editing.

% In contrast to previous methods, we construct image triplets using a real-world segmentation dataset, OpenImages\cite{kuznetsova2020open}, due to its large scale, diversity, and accurate annotation. 

To tackle the above issues, we construct the image triplet by applying a twice-repainting scheme based on the existing real-world segmentation dataset, OpenImages~\cite{kuznetsova2020open}, due to its large scale, diversity, and accurate annotation.
As depicted in Figure~\ref{fig:data}a, given a source image $I_s$ in the segmentation dataset, its instance mask $M$ and category label $C$, we repaint the mask region $M$ of the source image $I_s$ with the textual description related to the category label $C$ as the input prompt to produce the generated image $I_g$.
Simultaneously, we repaint the background region $\overline{M}=1-M$ of the source image $I_s$ to produce reference image $I_r$.
This allows us to construct an image triplet ($I_g$, $I_r$, $I_s$) for the reference-based object replacement task.
To further expand the diversity of the data, we also repaint the $\overline{M}$ region of the generated image $I_g$ to obtain a new reference image $I_{r'}$, resulting in an additional image triplet ($I_s$, $I_{r'}$, $I_g$).

It is more challenging to construct the triplet for the reference-based object addition task, primarily because it is difficult to pre-determine the exact position of the object to be added within the original image. While adding a specific reference object to a background image is complex, removing an existing object is comparatively simpler. This process is similar to our implementation in the object replacement task, with a crucial difference: when repainting the masked area, the goal is to remove the object, not modify it.
In addition, given the versatility of our \ours{} that it can be used for plain-text instruction-driven editing, we also maintain the flipping task of object addition, i.e. object removal, in our dataset.

During the image triplet construction, we empirically employ BrushNet~\cite{ju2024brushnet} as the repainting model, and for object removal, we utilize \design{}~\cite{jia2024designedit}. When constructing the reference image, we apply horizontal flipping and affine transformations to both the foreground image and the corresponding mask. These augmentation techniques enhance the complexity and diversity of our dataset, thereby increasing its effectiveness for training.

\begin{figure*}[!t]
\centering
\includegraphics[width=7.1in]{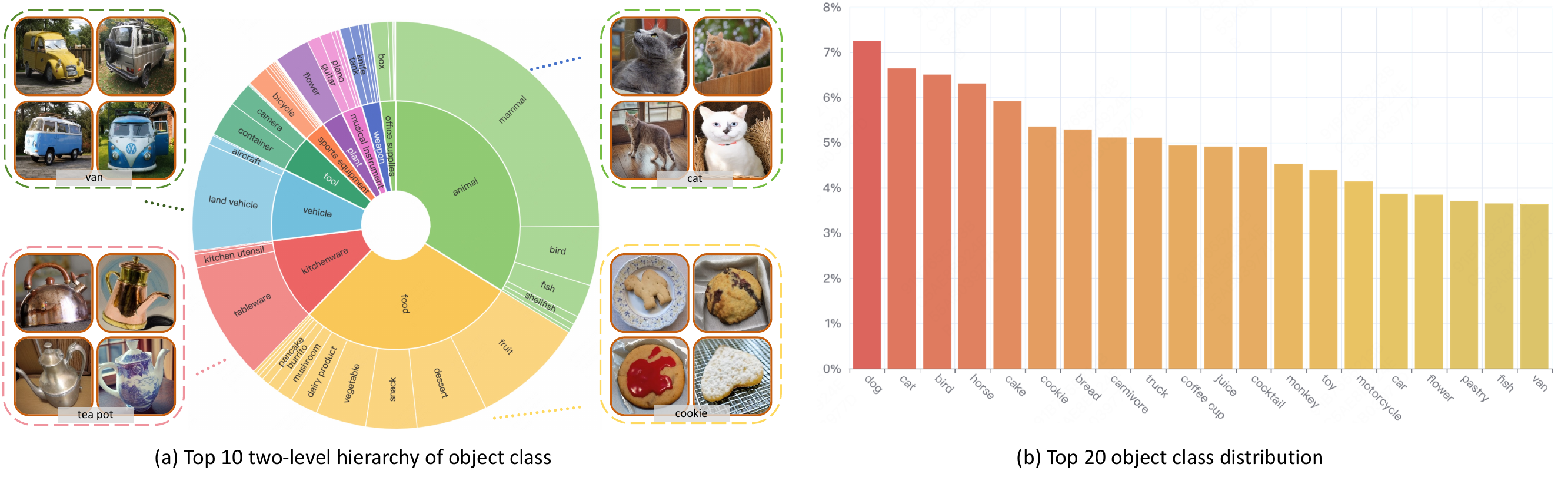}
\caption{Statistics for the \ourdata{} dataset. The first four parent classes in \ourdata{} are animals, food, kitchenware, and vehicles. \ourdata{} covers the vast majority of categories in daily life, allowing us to train a generalizable zero-shot reference-based image editing model.}
% Left: Top 10 two-level hierarchy of object class. Right: Top 20 object class distribution
% \color{red}{todo} Distribution statistics visualization of the category labels of the editing object in \ourdata{}. 
% 
\label{fig:category}

\end{figure*}

\begin{table}[!t]
\caption{Comparison of \ourdata{} with existing instruction-driven image editing datasets, including \ip2p{} and \mb{}.}
\label{tab:bench_comp}
\centering
\begin{tabular}{c | c c c c}
\hline
\textbf{Datasets} & \makecell[c]{{\bf Real Image} \\ {\bf Based?}} & \makecell[c]{{\bf Automatic} \\ {\bf Generated?}} & \makecell[c]{{\bf Multi-} \\ {\bf modal?}} & \textbf{\#Edits}\\
\hline 
\makecell[c]{\ip2p{}} & \makecell[c]{\cuo{}} & \makecell[c]{\dui{}} & \makecell[c]{\cuo{}} & 313,010 \\
% \makecell[c]{\mb{}~\citep{Zhang2023MagicBrush}} & \makecell[c]{\dui{}} & \makecell[c]{\cuo{}} & \makecell[c]{\cuo{}} & 10,388 \\
\makecell[c]{\mb{}} & \makecell[c]{\dui{}} & \makecell[c]{\cuo{}} & \makecell[c]{\cuo{}} & 10,388 \\
\ourdata{} (ours)    & \makecell[c]{\dui{}} & \makecell[c]{\dui{}} & \makecell[c]{\dui{}} & 131,160 \\
\hline
\end{tabular}

\end{table}

\subsection{Instruction Construction}
\label{data:ins}
To effectively support instruction-driven editing tasks, it is crucial to obtain editing instructions for each image triplet that are as detailed and accurate as possible. A straightforward approach might be to populate task-specific templates with category labels inherited from the segmentation dataset, thereby generating basic editing instructions. However, relying solely on category labels to refer to the object to be edited is overly simplistic. This approach can diminish the model's capacity to comprehend more complex editing instructions, which is essential for handling sophisticated editing tasks.

Traditional non-reference instruction-driven editing datasets acquire instructions from LLMs concurrently with generating image description pairs. While we cannot obtain instructions in advance as they do, we can leverage Multi-modal Large Language Models (MLLMs)~\cite{wang2023cogvlm,alayrac2022flamingo,hu2024minicpm,zhu2023minigpt} to caption the edited area of the pre-edit image. This approach allows us to produce a detailed local description of the object or area that is intended for editing.

Specifically, we take multiple MLLMs to improve the diversity of instruction contents, including (1) CogVLM~\cite{wang2023cogvlm} with grounding capabilities, (2) \llava{}~\cite{liu2023llava} with \alphaclip{}~\cite{sun2023alphaclip} and (3) \mini{}~\cite{hu2024minicpm}.
The first two models can provide a holistic view when captioning objects; however, they often face hallucination issues, leading to incorrect captions. Therefore, we introduce the third MLLM, \mini{}, which only observes cropped foreground images, to address this issue. While it doesn't account for object location within the image, it focuses on the description of the object's appearance.
We select the final captions by calculating the CLIP similarity between the image and captions obtained by different means.

% We have observed that the descriptions generated by directly feeding the foreground of the object to be described into the MLLM are more accurate and less susceptible to hallucination issues compared to those produced by the grounding-enabled MLLMs. As illustrated in Figure~\ref{fig:data}b, we input the masked original image into the MLLM and prompt it with instructions like “describe the foreground area” to obtain a visual description of the object slated for editing. For the object addition task, the requirement shifts to obtaining a description of the scene surrounding the edit area. We adopt a similar approach, but instead of the masked foreground image, we feed the MLLM with the image patch of the edit area. The local descriptions generated by the MLLM are then integrated with our hand-crafted editing templates, as depicted in Figure~\ref{fig:data}b, to create the final editing instructions. We compared \cog{} \cite{wang2023cogvlm}, \mini{} \cite{hu2024minicpm}, and \llava{} against \alphaclip{} \cite{sun2023alphaclip}, and selected the lightweight \mini{} model for our instruction construction process due to its efficiency and effectiveness.

\subsection{Data Filtering}
\label{data:filter}

Given the limitations of our repainting models, there's a risk of inadequate edits or poor reference images, and the MLLMs might not always generate precise local descriptions. To mitigate this, we use data filtering to remove low-quality entries and ensure the integrity of our training data.

Firstly, we assess the similarity between pre- and post-editing images, recognizing that a high degree of similarity can be problematic, as it may render the edited and original images nearly indistinguishable. Therefore, we employ filters to exclude data with high similarity scores using both the image encoders of CLIP~\cite{Radford2021CLIP} and DINO~\cite{Caron2021DINO,Oquab2023DINOv2}. 
Furthermore, there remains a risk of inadvertently introducing unrelated objects into the background and taking over the reference image. To counteract this, we utilize CLIP and DINO image similarity metrics to identify and exclude instances where the edited result significantly deviates from the reference image.
To ensure the accuracy of the instructions, we also filter out the edits that have a low similarity between the foreground area of the images to be edited and the MLLM-generated local descriptions.
% A low similarity score is indicative of a description that is not sufficiently accurate. By employing this metric, we can effectively identify and exclude descriptions that may compromise the quality of the instructions used in our dataset.

During the data filtering stage, we filtered out about 10\% of the data, resulting in the retention of 70,634 edits for object replacement, 30,263 for object addition, and 30,263 for object removal tasks. This brings the total number of edits in our dataset, \ourdata{}, to 131,160. A comparison of \ourdata{} with existing instruction-driven datasets is presented in Table~\ref{tab:bench_comp}. Notably, \ourdata{} stands out as the exclusive dataset that supports multi-modal instructions.
% \sout{It is built upon real images and can be automatically generated, which allows for substantial scalability.}
We visualize the distribution of category labels in the filtered dataset in Figure~\ref{fig:category}, showcasing that our dataset encompasses a wide range of categories representative of everyday life.

\section{EXPERIMENTS}
\label{sec:exp}

\begin{figure*}[!t]
\centering
\includegraphics[width=7.2in]{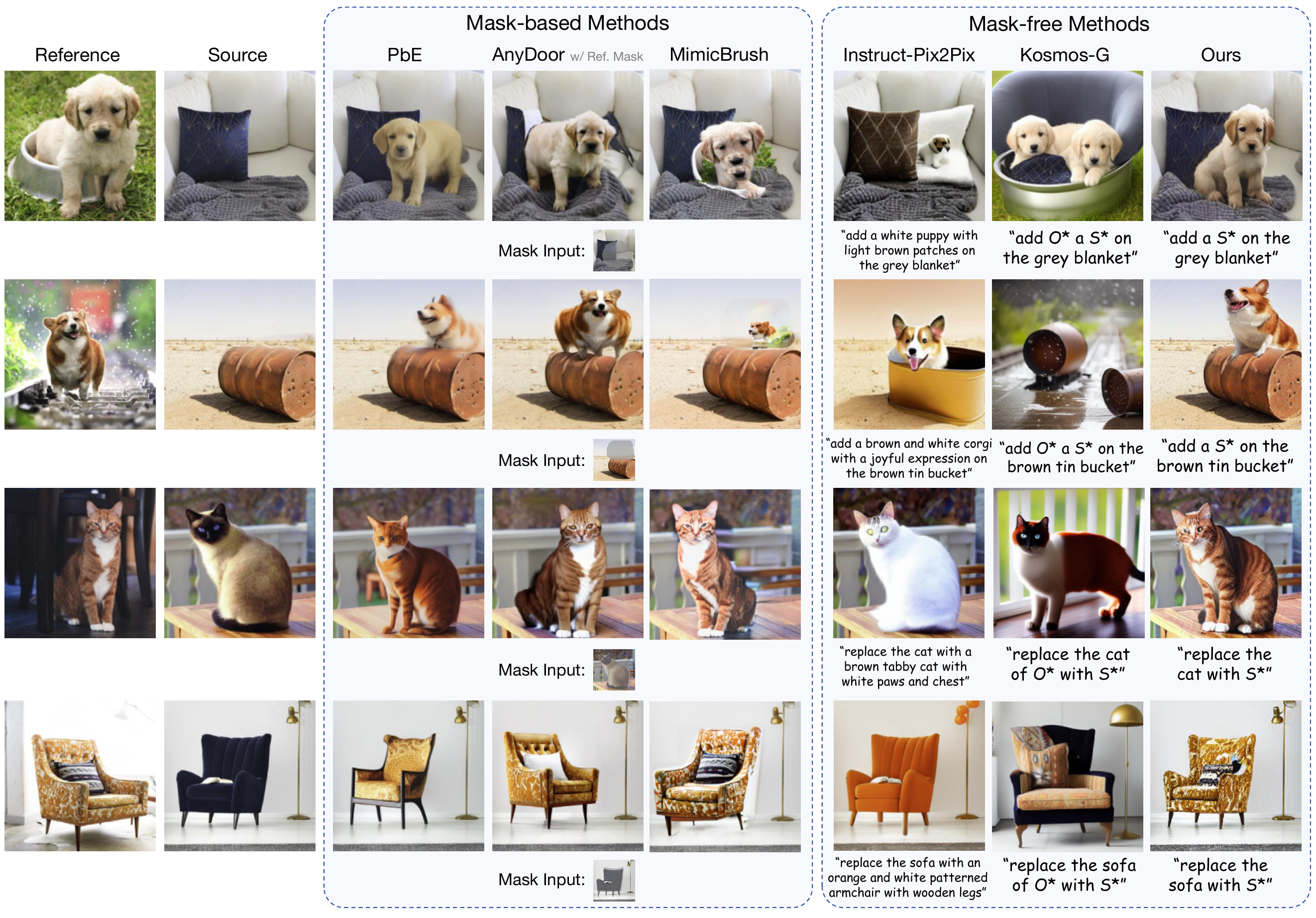}
\caption{Qualitative comparisons of \ours{} with previous methods, including mask-based methods \paintby{}, \anydoor{}, \mimic{}, and mask-free methods \ip2p{}, \kosmosg{}. Mask-based methods require the user to manually provide the mask of the editing area.
\anydoor{} also needs to provide a foreground mask of the reference image, and the editing mask fed to \anydoor{} will be processed as a box because its training is box-based.
Mask-free methods are language-based and don't require additional mask input, where we take \ip2p{} with detailed instructions as a baseline for comparison.
The inputs required for each method are marked below each line of images. S* denotes the specific visual concept in the reference image, and O* denotes the original image to be edited.}
\label{fig:comp}

\end{figure*}

\subsection{Implementation Details}

\noindent \textbf{Hyperparameters.}
We use Stable Diffusion V1-5\cite{von-platen-etal-2022-diffusers} in our experiments and modify its input channel from 4 to 8 to accommodate additional original image conditions.
The detail extractor, which shares the same architecture as Stable Diffusion V1-5, remains unchanged during training.
Our \qf{} is built on top of \blipdf{} which has 16 learnable queries.
Our experiments use 8 × 80GB NVIDIA A100 GPUs with AdamW optimizer, and the learning rate is set to 5e-6, with a total batch size of 512.
The \mm{} instruction training stage takes 20,000 steps, the fine-grained feature injection stage takes 5,000 steps, and the final quality-tuning stage takes 10,000 steps.
We train the model with the image at 256 × 256 resolution following \ip2p{}, and the trained model can be easily extended to 512 during inference.
During inference, we use 50 steps of the DDIM sampler, with text guidance at 5 and original image guidance at 1.5.

\noindent \textbf{Baselines.}
We compare our \ours{} with both mask-based and mask-free methods on the task of reference-based image editing.
The mask-based methods also require additionally provided masks of the editing areas, including:
(1) \paintby{}~\cite{yang2022paintby}, which introduces projected CLIP image embeddings to substitute the original text embeddings.
(2) \anydoor{}~\cite{chen2023anydoor}, which uses a DINO image encoder and high-frequency map to promote the reconstruction of reference concepts. It also requires a mask for the given reference image during inference.
% (3) \mimic{}~\cite{mimicbrush}, which achieves a similar task, imitative editing, by reference imitation.
(3) \mimic{}~\cite{mimicbrush}, which implements a similar task to reference-based image editing by referencing imitation, i.e., imitative editing.

There are limited mask-free methods supporting the task, and we adopt the following baselines, including:
(1) \ip2p{}~\cite{brooks2023instructpix2pix}, which doesn't directly support reference-based image editing. We provide it with text instructions containing detailed descriptions as a baseline method.
(2) \kosmosg{}~\cite{pan2023kosmosg}, a model designed for multi-modal image generation using both image and text inputs, which can serve as an alternative for reference-based image editing.

\noindent \textbf{Benchmarks.}
The previous approaches\cite{chen2023anydoor,yang2022paintby,mimicbrush} do not treat reference-based image editing as a distinct task, performing evaluation in the form of inpainting. However, reference-based image editing covers the tasks of reference-based object addition and replacement, which cannot be adequately assessed using traditional inpainting evaluation methods.
Image editing, being more nuanced than inpainting, often requires a detailed consideration of the original image's structure to formulate the corresponding editing requirements.
Consequently, we manually craft a new test benchmark to support the reference-based image editing task, including both object replacement and addition.
This dataset comprises 200 editing cases: 100 addition edits and 100 replacement edits, spanning 89 subjects and 96 scenes from everyday life.
Since the object removal task does not necessitate the use of reference images, it is not appropriate for joint evaluation with the aforementioned tasks. Therefore, we select an additional 100 object edits to serve as the test benchmark for the object removal task.
For methods that require masks, we also provide the corresponding editing masks to facilitate comparative analysis.

\noindent \textbf{Evaluation metrics.}
We employ L1 and L2 distance to measure the average pixel-level difference between the original images and edited images. CLIP-I and DINO-I are introduced to measure the similarity of the editing results to the original images, and CLIP-R and DINO-R are introduced to measure the similarity of the editing results to the reference images. CLIP-T is used to measure the similarity of the edited images to the edited text descriptions. We use ViT-B/32 for the CLIP image encoder and ViT-S/16 for the DINO image encoder in our experiments. 
Furthermore, to supplement these objective measurements, we also conduct user studies to gather subjective feedback. We distributed 50 questionnaires for both reference-based image editing and object removal tasks, and participants were asked to select the best edit from randomly arranged editing samples from different methods.

\subsection{Comparisons with Existing Methods.}

\begin{table*}[!t]
\caption{Quantitative comparison of \ours{} with existing methods on the reference-based image editing task.}
\label{tab:comp}
\centering
\begin{tabular}{l | c c c c c c c}
\toprule
Method & L1 $\downarrow$ & L2 $\downarrow$ & CLIP-I $\uparrow$ & DINO-I $\uparrow$ & CLIP-T $\uparrow$ & CLIP-R $\uparrow$ & DINO-R $\uparrow$\\
\midrule
\textbf{\textit{Mask-based Methods:}}\\
\paintby{} & 0.1028 & 0.0412 & 79.89 & 52.33 & 30.26 & \underline{76.73} & 48.39\\
\anydoor{} & 0.1108 & \underline{0.0470} & 78.46 & 48.62 & \textbf{30.79} & 77.88 & \textbf{60.72}\\
\mimic{} & \textbf{0.0772} & 0.0353 & \textbf{81.59} & \underline{54.78} & 29.73 & 74.99 & 53.83\\
\midrule
\textbf{\textit{Mask-free Methods:}}\\
\ip2p{} & 0.1908 & 0.0786 & 75.40 & 45.65 & \underline{30.58} & 57.23 & 5.87\\
\kosmosg{} & 0.2739 & 0.1185 & 74.42 & 41.77 & 28.05 & 76.65 & 50.26\\
\rowcolor{gray!10}\ours{} (ours)    & \underline{0.0806} & \textbf{0.0286} & \textbf{81.59} & \textbf{55.16} & 30.34 & \textbf{78.12} & \underline{54.37}\\
\bottomrule
\end{tabular}

\end{table*}

\noindent \textbf{Qualitative results.}
We provide qualitative results of reference-based image editing in Figure~\ref{fig:comp}, comparing our \ours{} with other baseline methods, encompassing both mask-based and mask-free approaches.
Generally, mask-based methods have an inherent advantage due to the additional mask input, as they do not need to infer the editing area from language instructions.
Still, their editing results are not ideal enough.
As an early reference-based inpainting work, \paintby{} results in edited images that only achieve a general resemblance in terms of color and shape.
\anydoor{} could recover more textures from the reference object, such as the orange-brown cat in the 3rd row.
Our concurrent work \mimic{}, introduces the reference imitation to retrieve intricate details from reference images. However, it is hard to apply to object-level editing tasks, as it occasionally leads to the presence of artifacts around the edited regions.

\begin{figure}[!t]
\centering
\includegraphics[width=3.5in]{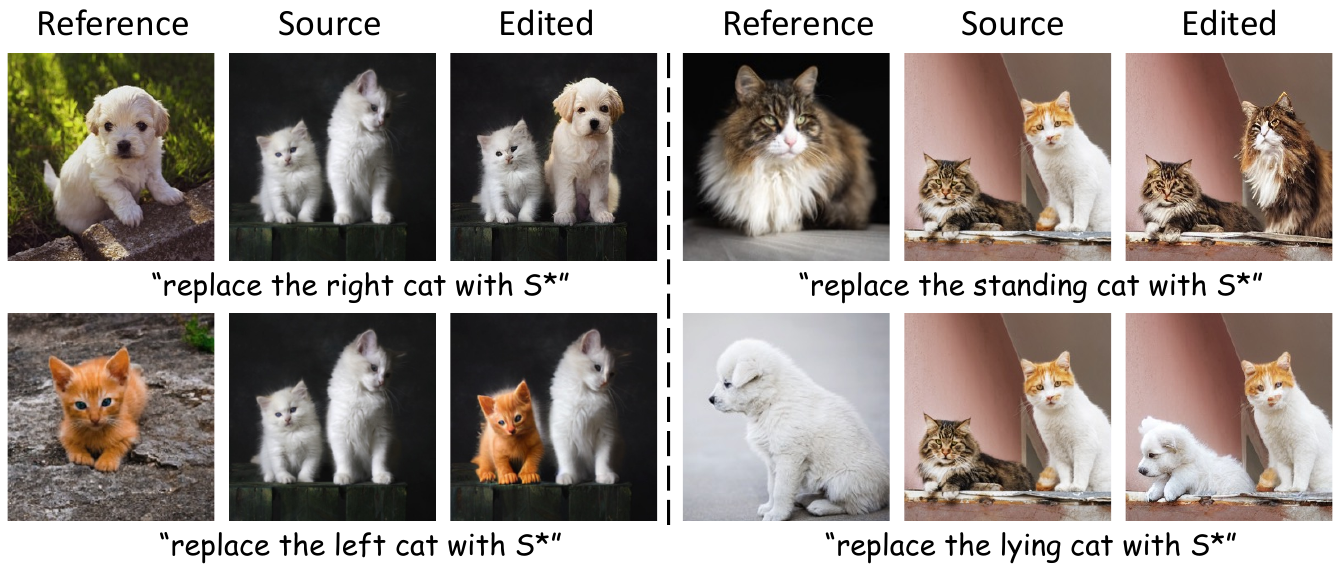}
\caption{\ours{} uses language instructions for convenient editing control without the introduction of manual masks, also capable of doing it when there are multiple objects in the original images.}
\label{fig:multi}
\end{figure}

The performance of previous mask-free methods is even worse due to the lack of explicit mask conditions and specialized reference feature encoding modules.
\ip2p{}, despite employing verbose text instructions, often fails to accurately restore the details of the reference subjects and leads to misinterpretations of the instructions, culminating in sub-optimal editing outcomes. For instance, as depicted in the 3rd row of Figure~\ref{fig:comp}, \ip2p{} erroneously replaces a cat with a white one, potentially due to the mention of white paws and chest in the instruction.
\kosmosg{}, while capable of multi-modal generation from text and images, still falls short in comparison to mask-based methods when it comes to the task of reference-based image editing.
In contrast, our \ours{} greatly improves the capabilities of the mask-free editing paradigm, surpassing even those methods that necessitate the use of masks.
Also shown in Figure~\ref{fig:multi}, it can locate the exact editing region based on the multi-modal instruction and yield harmonious editing outcomes that are closely aligned with the conceptual intent of the reference image.

\begin{figure}[!t]
\centering
\includegraphics[width=3.5in]{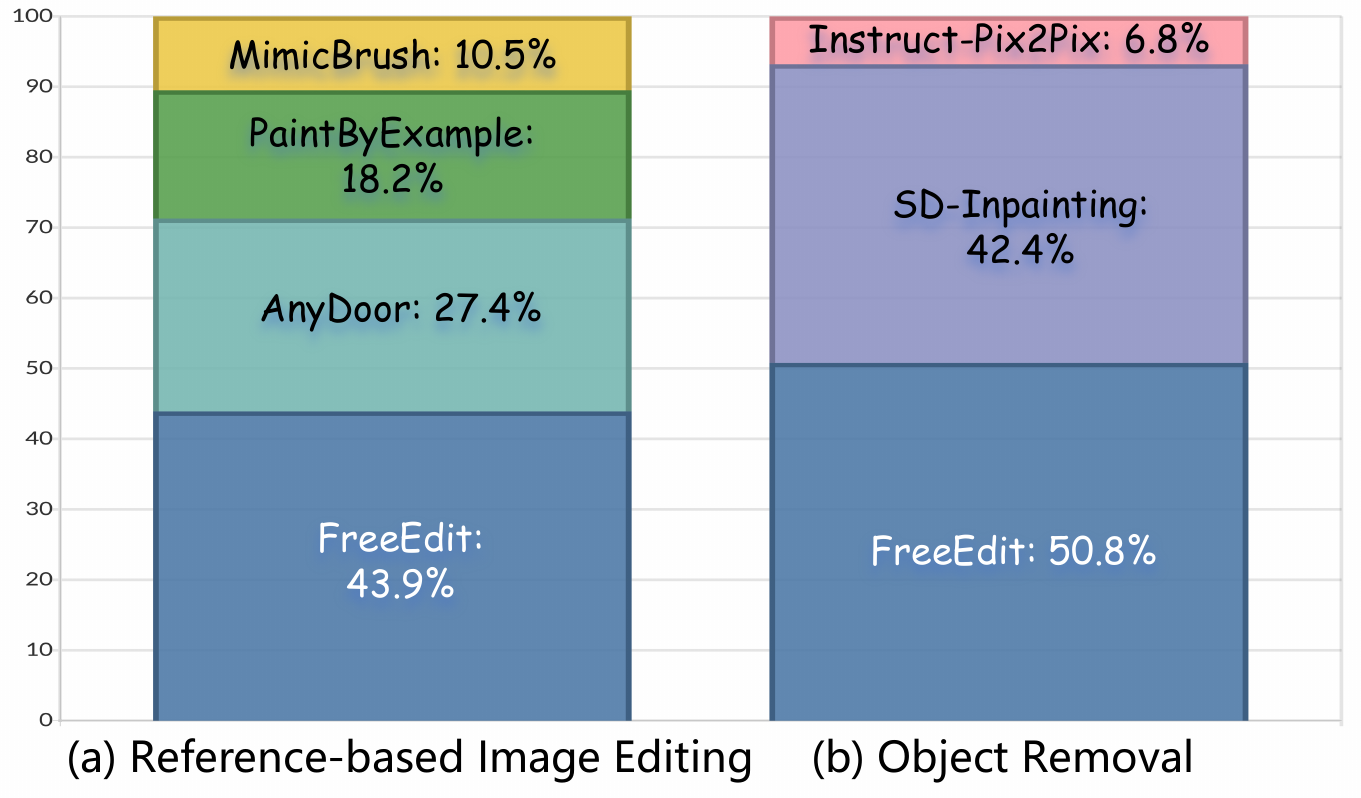}
\caption{User study results. Left: User study for the reference-based image editing task, we compare \ours{} with existing mask-based methods \anydoor{}~\cite{chen2023anydoor}, \paintby{}~\cite{yang2022paintby}, \mimic{}~\cite{mimicbrush}. Right: User study for the object removal task, we compare \ours{} with \ip2p{}~\cite{brooks2023instructpix2pix} and SD-Inpainting~\cite{Rombach2022StableDiffusion}.}
\label{fig:user}

\end{figure}

\noindent \textbf{Quantitative results.}
We present a quantitative comparison with other methods on the reference-based image editing task in Table~\ref{tab:comp}. 
Overall, mask-based methods have better similarities to the original image than previous mask-free methods, because the mask of the area to be edited has already been given in advance.
Our \ours{} significantly outperforms other mask-free methods across most metrics, achieving the highest original image similarity at CLIP-I and DINO-I, as well as the best reference image similarity at CLIP-R and DINO-R.
Even when compared to the mask-based methods, \ours{} is not lagging.
\ours{} is basically the same as \mimic{} in maintaining the similarity of the original image. In terms of maintaining the similarity of the reference image, \ours{} achieves the best CLIP-R, and second only to \anydoor{} in the DINO-R which achieves higher DINO-R scores due to the incorporation of DINO features during training.
The user study of reference-based image editing task illustrated in Figure~\ref{fig:user} shows that our \ours{} gains more favor from users.

% Additionally, given that existing quantitative metrics are not sufficient to measure relatively subjective editing effectiveness, we conducted a user study. Specifically, we distributed 50 questionnaires for both reference-based image editing and object removal, asking participants to choose the best ones from randomly shuffled options. 
% The results of the user study are shown in Figure~\ref{fig:user}, and our \ours{} gains more favor from users.

\begin{table}[!t]
\caption{Quantitative ablation study on the diversity of editing types.}
\label{tab:add_task}
\centering
\begin{tabular}{ccc | c c c c}
\toprule
Add & Replace & Remove & CLIP-T $\uparrow$ & CLIP-R $\uparrow$ & CLIP-IQA $\uparrow$\\
\midrule
\ding{51} &  &                    &30.00 & 75.97 &0.652\\
\ding{51} & \ding{51} &           &30.21 & 77.02 &0.655\\
\ding{51} & \ding{51} & \ding{51} &\textbf{30.44} & \textbf{77.15} &\textbf{0.705}\\
\bottomrule
\end{tabular}

\end{table}

\begin{table}[!t]
\caption{Ablation study about the training strategy. Recaptioning refers to recaptioning instructions, \qf{} refers to unfreezing \qf{} in training, DRRA refers to Decoupled Residual Refer-Attention.\label{tab:table1}}
\label{tab:big_abla}
\centering
\begin{tabular}{l | cccc}
\toprule
Method & CLIP-I $\uparrow$ & DINO-I $\uparrow$  & CLIP-R $\uparrow$ & DINO-R $\uparrow$\\
\midrule
Baseline &77.96 &51.54  &76.35 &49.44\\
+Recaptioning &77.95 &53.38  &76.63 &48.74\\
+\qf{} &75.17 &48.23  &76.63 &51.26\\
+DRRA &76.95 &52.29  &77.07 &\textbf{56.71}\\
+Quality Tuning &\textbf{81.59} &\textbf{55.16}  &\textbf{78.12} &54.37\\
\bottomrule
\end{tabular}
\end{table}

% \begin{table}[!t]
% \caption{Quantitative ablation study on refer-attention module.}
% \centering
% \label{tab:refer_attn}
% \begin{tabular}{l | c c c c c}
% \toprule
% Method & CLIP-I $\uparrow$ & DINO-I $\uparrow$ & CLIP-R $\uparrow$ & DINO-R $\uparrow$\\
% \midrule
% w/o RA &75.17 &48.23  &76.63 &51.26\\% &13.39\\
% RASA   &76.88 &51.77  &76.80 &56.29\\% &14.67\\
% DRRA   &\textbf{76.95} &\textbf{52.29}  &\textbf{77.07} &\textbf{56.71}\\% &14.92\\
% \bottomrule
% \end{tabular}
% 
% \end{table}

\subsection{Ablation Study}
\label{exp:abla}

% \textcolor{red}{Divide into two parts: dataset and method.}
In this section, we perform a series of ablation studies to assess the effectiveness of various components and strategies.
Firstly, for the dataset, we verify the importance of the diversity of editing types and instruction recaptioning. Secondly, in terms of model training, we verify the impact of training \qf{}, refer-attention module, and quality tuning. Finally, we perform an ablation experiment on the reference scale $\lambda$ in inference.

\noindent \textbf{Diversity of Editing Types.}
We generate three types of data during the dataset construction phase, including object addition, object replacement, and object removal. The first two types of edits are accompanied by reference images. To explore the influence of task types on model training, we conduct an ablation experiment. Specifically, we consider the reference object addition task, and the baseline is using only the Add type data in training. As shown in Table~\ref{tab:add_task}, when we append the Replace type data in the training, we find that CLIP-R is improved due to the introduction of more diverse reference tasks. Subsequently, we include the Remove type data, which does not involve a reference image, but still helps the model learn the instruction-driven editing behavior, resulting in an improvement in the image quality metric CLIP-IQA~\cite{wang2022clipiqa}. This indicates that incorporating various task types enhances the model’s editing capabilities.

\begin{figure*}[!t]
\centering
\includegraphics[width=7.2in]{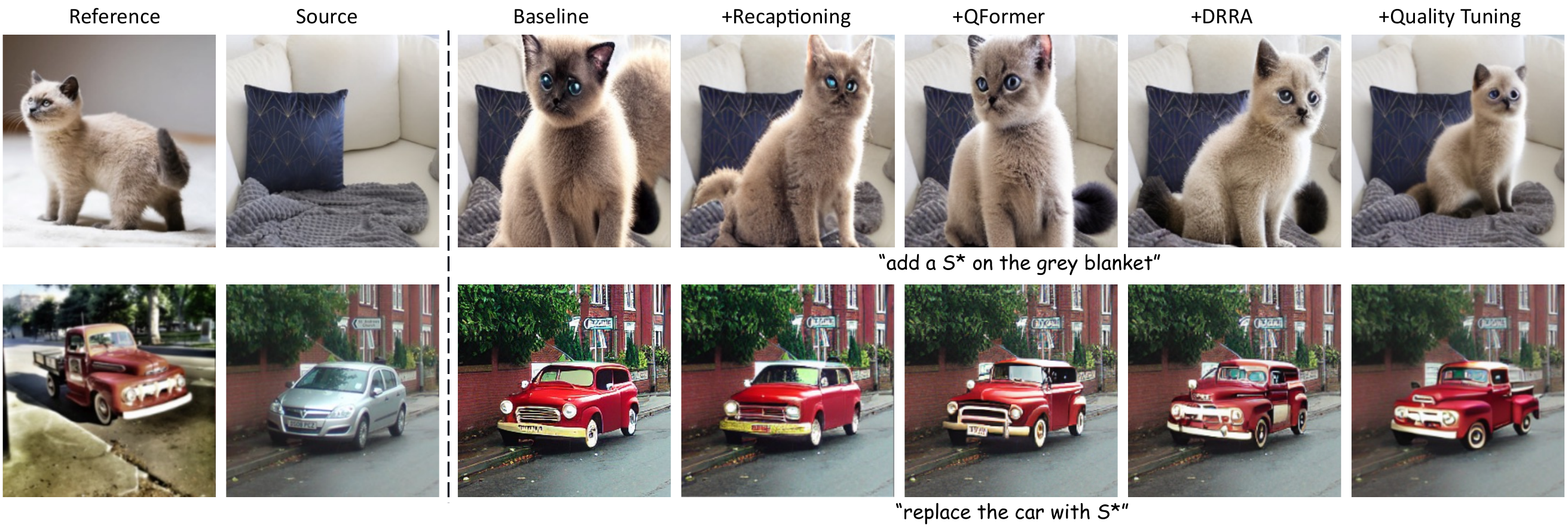}
\caption{Visual ablation studies of individual components in \ours{}. We gradually improve the quality of the editing and the similarity to the reference subject through these techniques, leading to promising reference-based instruction-driven editing results.}
\label{fig:abla_vis}

\end{figure*}

\noindent \textbf{Instruction Recaptioning.}
We conduct ablation experiments to validate the effectiveness of our instruction recaptioning. As described in Section~\ref{data:ins}, we use the MLLMs to recaption local areas of the images to be edited, which are then combined with the editing templates to produce more detailed and precise instructions. The baseline method is to populate the template with the category labels of the object to be edited directly, without involving the fine-grained references. As shown in Table~\ref{tab:big_abla}, instruction recaptioning helps to improve the consistency of images before and after editing, as evidenced by the improvement in DINO-I scores.
As shown in the 1st row of Figure~\ref{fig:abla_vis}, with the introduction of recaptioning, the shape of the cat in the edited image is more reasonable, leading to fewer artifacts.

\noindent \textbf{Training \qf{}.}
\qf{} is frozen in the previous work MiniGPT4~\cite{zhu2023minigpt}, as it has been trained to extract visual features conditioned on the reference text from the reference image. We conduct ablation experiments to verify the impact of training \qf{}. As shown in Table~\ref{tab:big_abla}, training \qf{} enables it to extract features that better align with the reference concept, leading to an improvement in DINO-R similarity to the subject in the reference image.
As shown in Figure~\ref{fig:abla_vis}, the editing results of training \qf{} are more consistent with the reference subjects.

\begin{figure}[!t]
\centering
\includegraphics[width=3.5in]{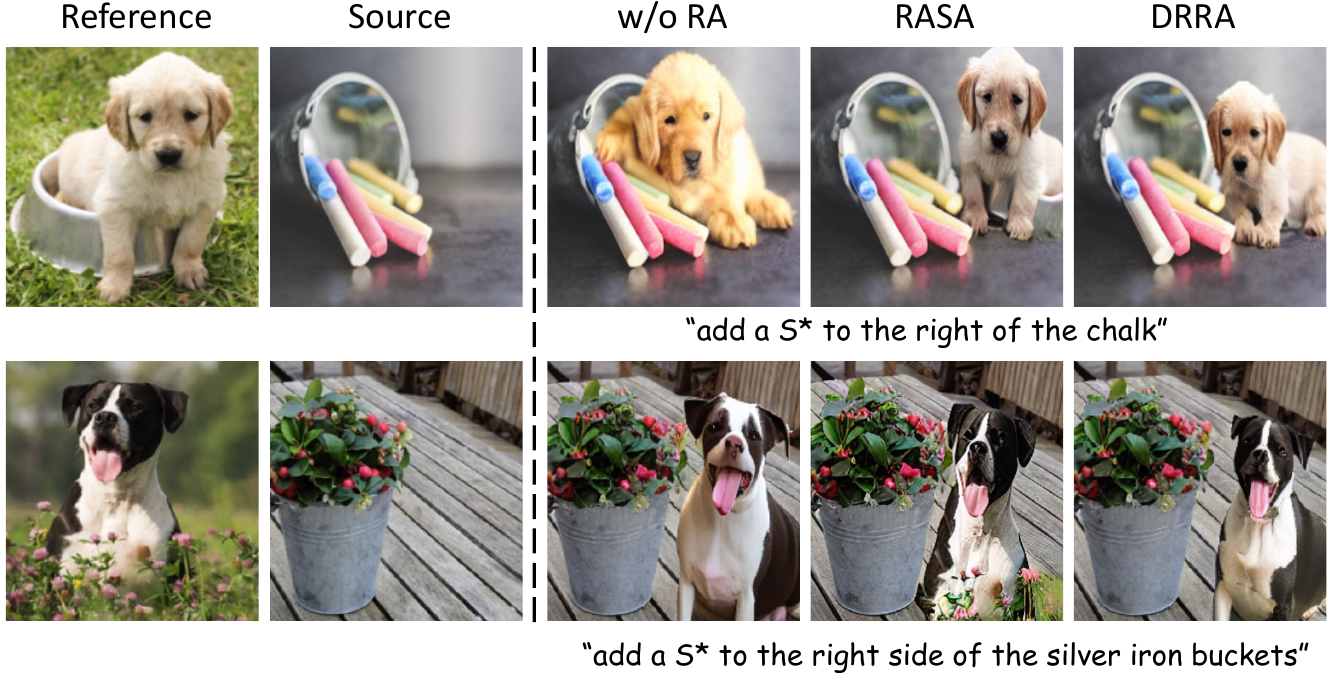}
\caption{Qualitative ablation study on the refer-attention modules. w/o RA. denotes the primitive self-attention module without the introduction of reference features. RASA denotes Refer-Attention in Self-Attention. DRRA denotes Decoupled Residual Refer-Attention.}
\label{fig:abla}

\end{figure}

\begin{table}[!t]
\caption{Quantitative ablation study on refer-attention module.}
\centering
\label{tab:refer_attn}
\begin{tabular}{l | c c c c c}
\toprule
Method & CLIP-I $\uparrow$ & DINO-I $\uparrow$ & CLIP-R $\uparrow$ & DINO-R $\uparrow$\\
\midrule
w/o RA &75.17 &48.23  &76.63 &51.26\\% &13.39\\
RASA   &76.88 &51.77  &76.80 &56.29\\% &14.67\\
DRRA   &\textbf{76.95} &\textbf{52.29}  &\textbf{77.07} &\textbf{56.71}\\% &14.92\\
\bottomrule
\end{tabular}

\end{table}

\begin{figure}[!t]
\centering
\includegraphics[width=3.5in]{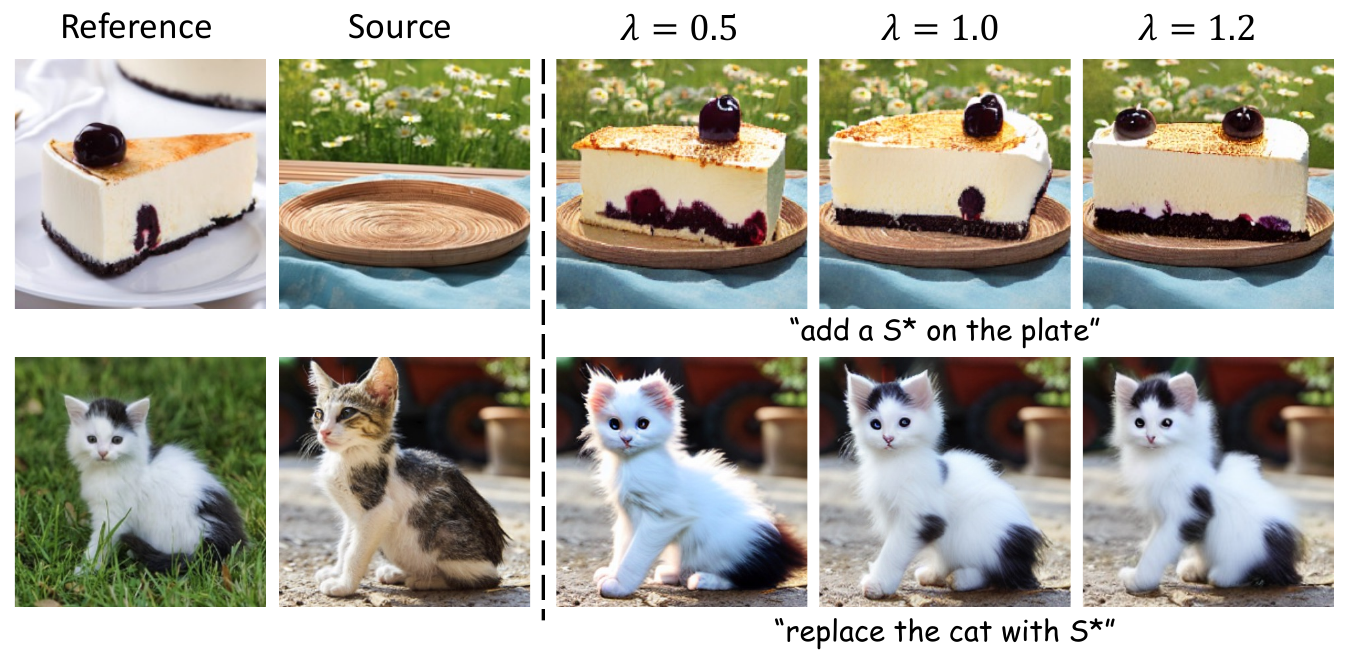}
\caption{Effect of hyperparameter scale $\lambda$. A larger $\lambda$ makes the generated region more similar to the reference.}
\label{fig:scale}

\end{figure}

\noindent \textbf{Refer-Attention.}
We conduct ablation experiments to validate the effectiveness of our newly proposed DRRA, which uses decoupled refer-attention to introduce detailed features while preserving the original self-attention mechanism.
The baseline methods include naive self-attention, which lacks reference attention, denoted as w/o RA, and Refer-Attention in Self-Attention, denoted as RASA.
The quantitative comparisons are presented in Table~\ref{tab:refer_attn}. Both DRRA and RASA improve CLIP-R and DINO-R, resulting in better ID consistency with reference subjects. Among them, DRRA outperforms RASA on CLIP-I and DINO-I metrics. 
Figure~\ref{fig:refer_attn} provides detailed visual comparisons of the different components.
Naive self-attention struggles to reconstruct specific concepts in reference images.
DRRA maintains a high degree of similarity between the generated subjects and the ones in the reference images while introducing fewer extraneous background features compared to RASA, leading to superior editing quality.
It can also be observed in Figure~\ref{fig:abla_vis} that the introduction of DRRA produces editing results that are more consistent with the characteristics of the reference subjects.

\noindent \textbf{Quality Tuning.}
To further enhance the performance of image editing, we conduct a quality tuning stage on the trained model. To verify the effectiveness of the quality tuning, we perform a corresponding ablation experiment. As shown in Table~\ref{tab:big_abla}, the quality-tuned model shows a better balance between the similarity of the original image and the similarity of the reference image, achieving the best results on the CLIP-I, DINO-I, and CLIP-R indicators, which underscores the important role of the quality tuning stage. This is also consistent with the observations in our visual ablation experiment shown in Figure~\ref{fig:abla_vis}. Compared to the editing results without quality tuning, the generated cat in the 1st row is more harmonious with the original image background, and the replacement of the car in the 2nd row also has better image quality.

\noindent \textbf{Reference Scale.}
As mentioned in Section~\ref{method:kv}, one of our advantages over the previous methods is that we can flexibly control the degree of attention paid to the reference image in the editing process through the hyper-parameter $\lambda$. As shown in Figure~\ref{fig:scale}, the edited objects become more and more similar to the reference images as $\lambda$ grows. In our experiment, we set $\lambda$ = 1 by default.

\subsection{Object Removal}
\label{exp:rm}

\begin{table}[!t]
\caption{Quantitative comparison of \ours{} with \ip2p{} and SD-Inpainting on the object removal task.}
\label{tab:remove_task}
\centering
\begin{tabular}{l | c c c c}
\toprule
Method & PSNR $\uparrow$ & DINO-I $\uparrow$ & CLIP-I $\uparrow$ & LPIPS $\downarrow$\\
\midrule
\ip2p{}       & 9.59 & 25.87& 61.52 & 55.41\\
SD-Inpainting & 17.75& 75.74& 87.27& 16.96\\
\ours{}(ours) & \textbf{19.01}& \textbf{77.01}& \textbf{90.03}& \textbf{14.80}\\
\bottomrule
\end{tabular}

\end{table}

\begin{figure}[!t]
\centering
\includegraphics[width=3.5in]{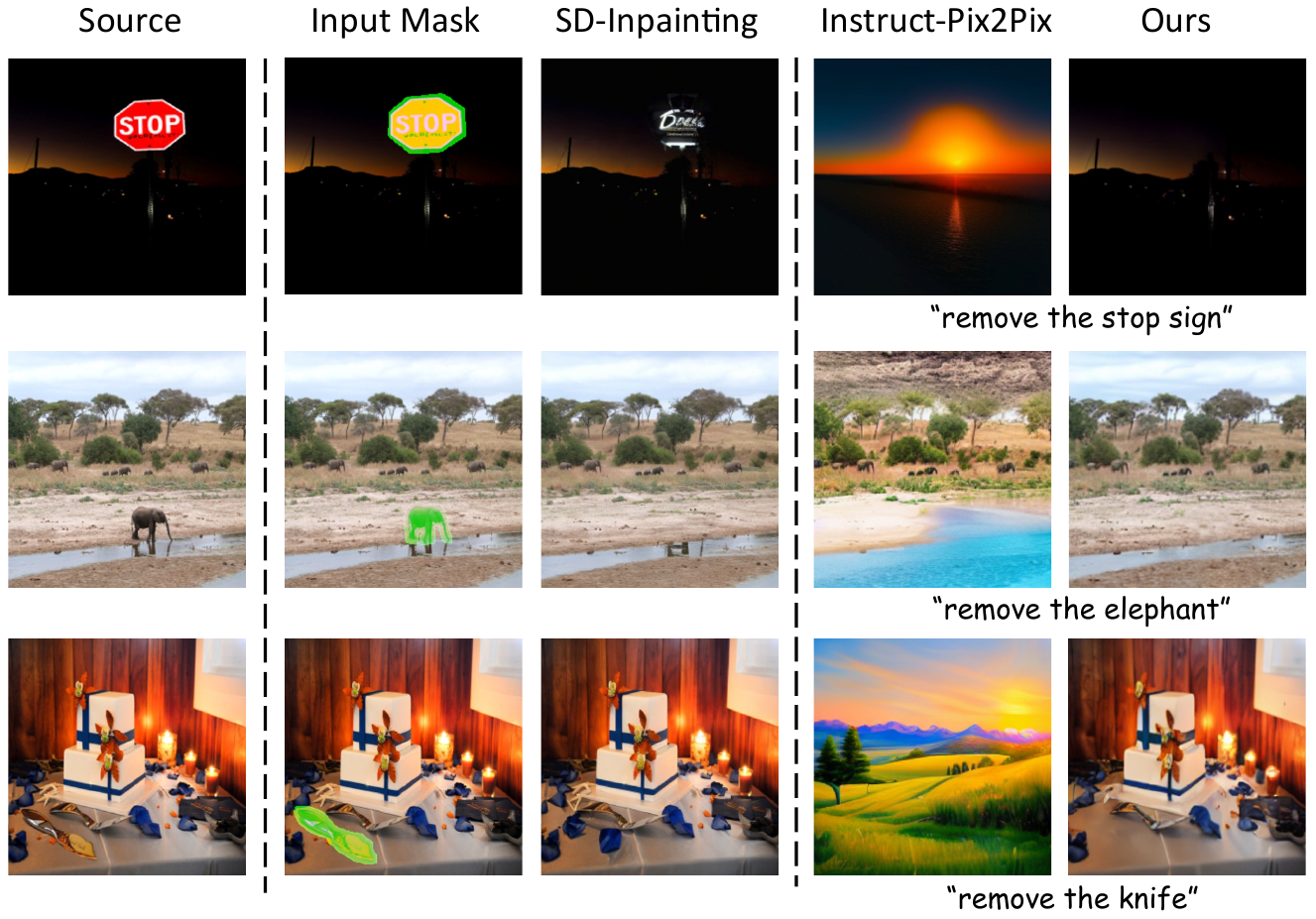}
\caption{A qualitative comparison of \ours{} with SD-Inpainting and \ip2p{} on the object removal task. SD-Inpainting requires the edit area mask of the object to be deleted, while our \ours{} and \ip2p{} perform object removal without the need for masks in the form of language instructions. The input mask for SD-Inpainting is highlighted in green in the 2nd column.}
\label{fig:task_rm}

\end{figure}

\ours{} also supports reference-free object removal task by simply setting the reference scale $\lambda$ to 0.
We compare the object removal capability of \ours{} with \ip2p{}~\cite{brooks2023instructpix2pix} and SD-Inpainting~\cite{Rombach2022StableDiffusion}.
\ours{} and \ip2p{} utilize editing instructions, whereas SD-Inpainting employs appropriate text descriptions with extra masks that specify the regions for editing.
The quantitative evaluation is carried out on a test set of 100 object removal cases which are synchronously collected when we construct the reference-based editing test set.

The quantitative comparison results are presented in Table~\ref{tab:remove_task}. SD-Inpainting and \ours{} are superior in terms of evaluation metrics compared to \ip2p{}.
This aligns with the visual evidence provided in Figure~\ref{fig:task_rm}, which indicates that \ip2p{} struggles to preserve the coherence with the original image during object removal.
\ours{} delivers results that are on par with SD-Inpainting while minimizing the introduction of artifacts.
For instance, in the 1st row, \ours{} effectively removes the stop sign without leaving noticeable traces. Furthermore, \ours{} successfully erases the reflection of the elephant on the river surface in the 2nd row, and in the 3rd row, it achieves a clean deletion without the unwanted creation of new objects, thus achieving better editing results compared with SD-Inpainting.
The user study of the object removal task illustrated in Figure~\ref{fig:user} also shows that our \ours{} gains more favor from users.

\subsection{Plain-text Instruction-driven Editing}

\begin{table}[!t]
\caption{Quantitative comparison of \ours{} with \ip2p{} and \mb{} on the plain-text instruction-driven editing task.}
\label{tab:text_task}
\centering
\begin{tabular}{l | c c c c}
\toprule
Method & CLIP-I $\uparrow$ & CLIP-T $\uparrow$ & CLIP-D $\uparrow$ & PSNR $\uparrow$ \\
\midrule
\ip2p{}       & 67.21         & 25.02         & 6.14          & 10.02\\
\mb{}         & 87.53         & 26.49         & \textbf{11.45}& \textbf{15.28}\\
\ours{}(ours) & \textbf{88.39}& \textbf{26.98}& 10.14         & 15.12\\
\bottomrule
\end{tabular}

\end{table}

\begin{figure}[!t]
\centering
\includegraphics[width=3.5in]{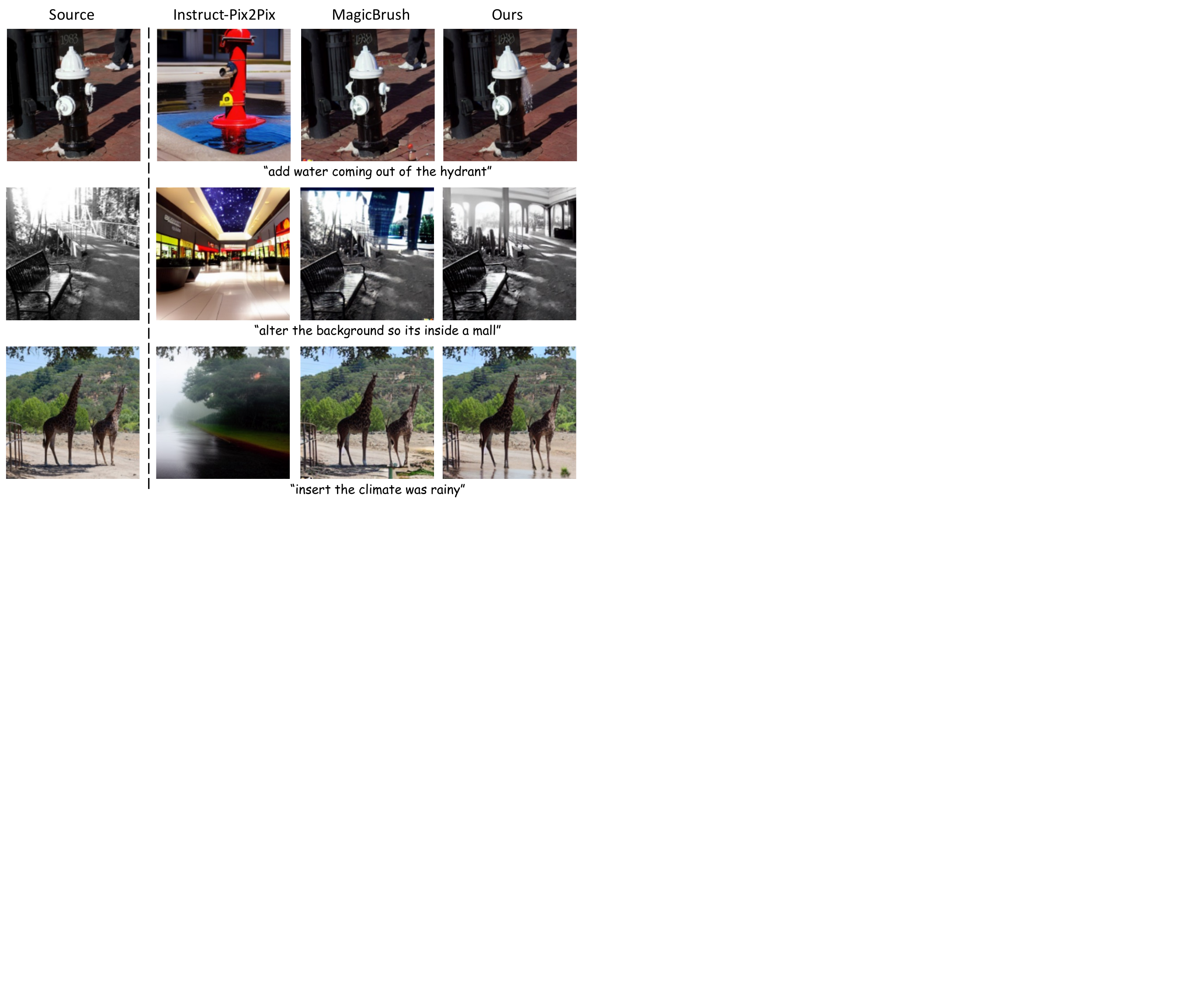}
\caption{The Visual comparison of \ours{} with \ip2p{} and \mb{} on plain-text instruction-driven editing task. The corresponding text instruction used for each edit case is marked below the corresponding images.}
\label{fig:comp_text}

\end{figure}

% As with object removal task, \ours{} also supports a broader range of plain-text instruction-driven editing, as our multi-modal instructions are closely related to text instructions.
As with object removal task, \ours{} also supports a broader range of plain-text instruction-driven editing which is not limited to a specific type of editing, as our multi-modal instructions are closely related to text instructions.
In this section, we assess this task from a comprehensive standpoint.
% Diverging from the conventional inpainting-based methods that condition the diffusion model on projected image embeddings, \ours{} incorporates the multi-modal instruction, which establishes a closer connection to the text space, enabling our model to not only perform reference-based editing but also to handle plain-text instruction-driven editing.
% The object removal task, as elaborated in Section~\ref{exp:rm}, can be regarded as a specialized instance of plain-text instruction-driven editing. In this section, we assess this task from a comprehensive standpoint.

We quantitatively compare \ours{} with previous methods \ip2p{} and \mb{} in the \emuedit{} test split, as shown in Table~\ref{tab:text_task}. \ours{} demonstrates the similarity with the original image and the faithfulness of the text instruction compared with \ip2p{}, and has comparable performance with \mb{}.
The visual comparisons are shown in Figure~\ref{fig:comp_text}. Compared to \ip2p{}, \mb{} and \ours{} could maintain the higher similarity between the editing result and the original image. Interestingly, we observe that \ours{} shows better text instruction fidelity than \mb{} in some cases, such as water gushing out of the hydrant in the 1st row, and the ground getting wet and showing the reflection of a giraffe in the 3rd row, which we speculate may be due to the diverse reference-based editing dataset which gives the model a more nuanced comprehension of editing instructions.

\subsection{Mask-Free Virtual Try-On}

\begin{figure}[!t]
\centering
\includegraphics[width=3.5in]{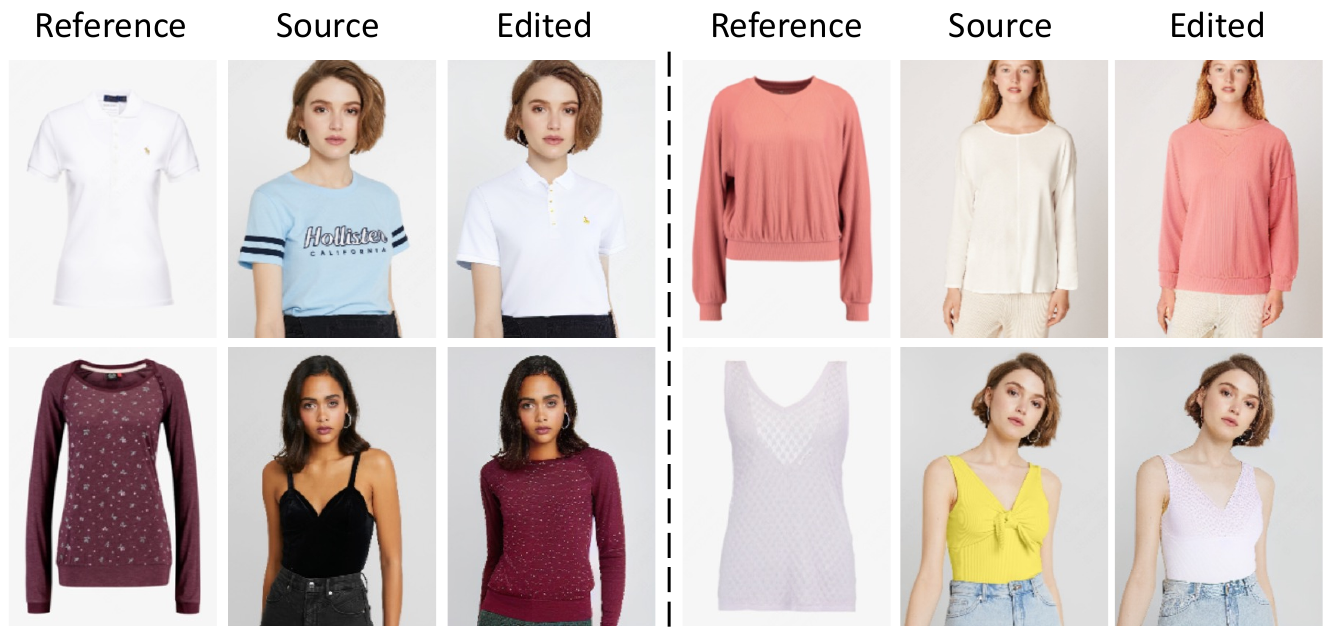}
\caption{Application of \ours{} on instruction-driven mask-free try-on on VitonHD-test\cite{choi2021viton}. The editing instruction is all “replace her top with S*”. Users can implement convenient virtual try-ons based on language instructions and corresponding reference images without the intervention of masks.}
\label{fig:tryon}

\end{figure}

As a versatile reference-based image editing model, \ours{} can be extended to virtual try-on task, as illustrated in Figure~\ref{fig:tryon}. Different from the previous methods, \ours{} simplifies the inference pipeline of virtual try-on. Users could execute the task according to a concise multi-modal language instruction that conforms to human habits and does not need to provide a manual mask.
Despite not being specifically trained on virtual try-on task data, \ours{} demonstrates significant promise in this domain.
Even when clothing depicted in the reference image significantly differs from that in the original image, such as the purple long-sleeve top shown in the last row of Figure~\ref{fig:tryon}, our model is capable of delivering a reasonable and visually coherent editing result.

\section{Conclusion}

In this paper, we present \ours{}, designed for the task of reference-based image editing in a mask-free manner.
To support model training, we build the first dataset supporting the task, \ourdata{}, through a newly developed twice-repainting scheme.
\ours{} contrasts with traditional inpainting-based methods by eliminating the requirement for users to define the editing region through masking.
Instead, \ours{} employs \mm{} instruction encoder to encode language instructions, which are more in line with human editing habits, to implicitly locate the editing area.
To overcome the limitations of \mm{} instructions in the high-fidelity reconstruction of reference concepts, we introduce the DRRA module. This module facilitates the seamless incorporation of the detailed reference features derived from a detail extractor into the image editing workflow.
By training in stages on \ourdata{} and then performing quality tuning, \ours{} could perform excellent reference-based image editing and also support plain-text editing including object removal.

\section{Limitation}

\ours{} significantly improves the performance of the reference-based image editing task in a mask-free manner.
At present, it only accommodates a single reference visual concept in an edit, yet future developments could potentially extend its capabilities to handle more intricate editing tasks involving multiple reference concepts.

\bibliography{IEEEabrv,ref}
\bibliographystyle{IEEEtran}

\end{document}